\documentclass[sigconf]{acmart}
\AtBeginDocument{%
  }

\setcopyright{acmlicensed}
\copyrightyear{2018}
\acmYear{2018}
\acmDOI{XXXXXXX.XXXXXXX}
\acmConference[]{}{}{}
\acmISBN{978-1-4503-XXXX-X/2018/06}

\usepackage{booktabs}      
\usepackage{multirow}      
\usepackage{graphicx}      
\usepackage{array}         
\usepackage{makecell}      
\usepackage{caption}       
\usepackage{subcaption}
\usepackage{adjustbox}
\usepackage{booktabs}
\usepackage{amsmath}

\usepackage{amssymb, amsmath, multirow, enumitem, colortbl, fancyhdr}
\definecolor{tabred}{rgb}{1, 0.7, 0.7}
\definecolor{taborange}{rgb}{1, 0.85, 0.7}
\definecolor{tabcapred}{rgb}{1, 0.65, 0.65}
\definecolor{tabcaporange}{rgb}{1, 0.7, 0.5}
\definecolor{figblue}{rgb}{0, 0.6902, 0.9412}
\definecolor{figorange}{rgb}{0.9294, 0.4902, 0.1921}

\renewcommand\footnotetextcopyrightpermission[1]{} 
\begin{document}
\fancypagestyle{standardpagestyle}{
  \fancyhf{}
  \renewcommand{\headrulewidth}{0pt}
  \renewcommand{\footrulewidth}{0pt}
}
\pagestyle{standardpagestyle} 
\title{Generate Aligned Anomaly: Region-Guided Few-Shot Anomaly Image-Mask Pair Synthesis for Industrial Inspection}

\author{Yilin Lu$^{1}$}
\authornote{Equal Contribution.}
\author{Jianghang Lin$^{1}$}
\authornotemark[1]
\author{Linhuang Xie$^{1}$} 
\author{Kai Zhao$^{2}$\\}
\author{Yansong Qu$^{1}$}

\author{Shengchuan Zhang$^{1}$}
\authornote{Corresponding author.}
\author{Liujuan Cao$^{1}$}
\author{Rongrong Ji$^{1}$}
\affiliation{%
\institution{$^{1}$ Key Laboratory of Multimedia Trusted Perception and Efficient Computing,\\ Ministry of Education of China, Xiamen University, China.~$^{2}$ VIVO.}
}

\begin{abstract}
Anomaly inspection plays a vital role in industrial manufacturing, but the scarcity of anomaly samples significantly limits the effectiveness of existing methods in tasks such as localization and classification. While several anomaly synthesis approaches have been introduced for data augmentation, they often struggle with low realism, inaccurate mask alignment, and poor generalization. To overcome these limitations, we propose Generate Aligned Anomaly (GAA), a region-guided, few-shot anomaly image–mask pair generation framework. GAA leverages the strong priors of a pretrained latent diffusion model to generate realistic, diverse, and semantically aligned anomalies using only a small number of samples. The framework first employs Localized Concept Decomposition to jointly model the semantic features and spatial information of anomalies, enabling flexible control over the type and location of anomalies. It then utilizes Adaptive Multi-Round Anomaly Clustering to perform fine-grained semantic clustering of anomaly concepts, thereby enhancing the consistency of anomaly representations. Subsequently, a region-guided mask generation strategy ensures precise alignment between anomalies and their corresponding masks, while a low-quality sample filtering module is introduced to further improve the overall quality of the generated samples. Extensive experiments on the MVTec AD and LOCO datasets demonstrate that GAA achieves superior performance in both anomaly synthesis quality and downstream tasks such as localization and classification.
\end{abstract}


\begin{teaserfigure}
  \includegraphics[width=\textwidth]{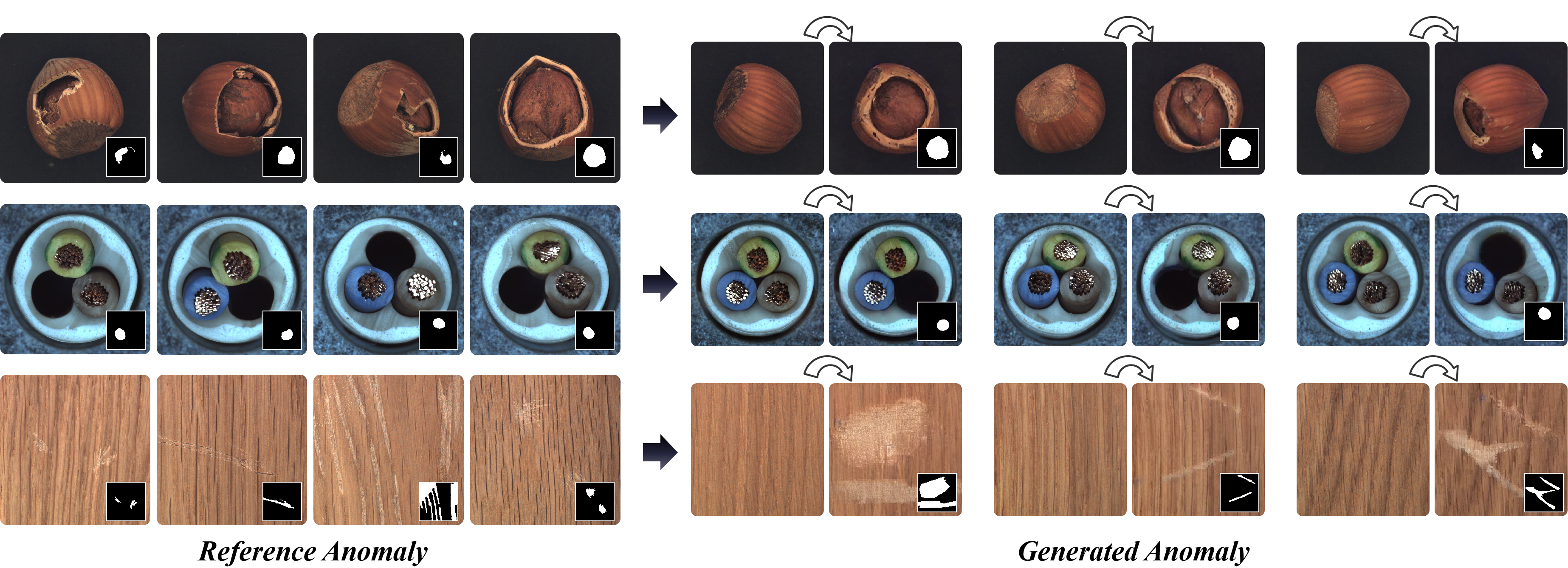}
  \caption{Given a few image–mask pairs, GAA learns anomaly concepts via a pretrained latent diffusion model and adaptively generates aligned masks to guide anomaly synthesis on normal images. It enables controllable generation of high-fidelity anomalies, including structural (e.g., cracks on hazelnuts, Row 1), logical (e.g., missing cables, Row 2), and compositional (e.g., scratches on wood, Row 3), supporting tasks such as anomaly detection, localization, and classification.}
  \label{fig:teaser}
\end{teaserfigure}


\maketitle
\pagestyle{empty}
\section{Introduction}
Anomaly inspection has become a core technique in modern industrial manufacturing to ensure product quality and production safety. This task involves multiple subtasks, including anomaly detection, localization, and classification, which place high demands on the accuracy and generalization ability of models. However, due to the extreme scarcity of anomaly samples in real-world industrial scenarios, most visual anomaly detection methods rely on unsupervised learning~\cite{roth2022towards, liu2023simplenet,chaudhry2018riemannian}, such as training one-class classifiers solely on normal data. Some approaches also attempt supervised learning with a few anomaly samples~\cite{jeong2023winclip,pang2021explainable}. Although these methods perform well in detection, they generally show limited performance in localization and are inadequate for anomaly classification. Therefore, addressing the shortage of anomaly data has become increasingly urgent.

Recently, anomaly data synthesis has emerged as a promising direction to address this issue by expanding the anomaly dataset and enhancing the robustness of supervised models in downstream tasks. Existing synthesis methods can be roughly categorized into two types:(1)~Model-free methods~\cite{li2021cutpaste, zavrtanik2021draem}: These methods directly paste anomaly textures or patches onto normal images. While simple and efficient, they often lack semantic plausibility and struggle to simulate realistic anomalies. (2)~Generation-based methods: These approaches leverage GANs or diffusion models to generate more natural-looking anomalies. For example, DiffAug~\cite{zhang2021defect} enhances diversity in few-shot settings, while SDGAN~\cite{niu2020defect} emphasizes structural consistency. However, most of these methods cannot simultaneously generate anomaly masks, limiting their applicability to localization tasks. DFMGAN~\cite{duan2023few} attempts to generate both anomalies and masks, but the lack of explicit spatial alignment mechanisms leads to limited realism.
A recent method, AnoDiff~\cite{hu2024anomalydiffusion}, proposes a two-stage strategy to generate both anomaly masks and images via diffusion models, offering a new perspective for mask-guided anomaly synthesis. Nevertheless, it still suffers from two key limitations: (1) the model does not explicitly distinguish between different types of anomalies, relying instead on mixed feature modeling, which results in semantically ambiguous and less realistic outputs; and (2) the generated masks often fall on background or irrelevant regions, causing misalignment between anomalies, masks, and object regions, which degrades realism and downstream performance.

Moreover, current anomaly synthesis paradigms exhibit limited generalization. Most existing methods are restricted to simulating structural surface anomalies on single objects, while performance significantly degrades when encountering more complex industrial anomalies, such as logical or compositional anomalies. In such cases, the generated samples often lack realism and diversity, making them easily distinguishable from real-world anomalies.

To address the above challenges, we propose \textbf{Generate Aligned Anomaly(GAA)}, a region-guided few-shot anomaly image–mask pair synthesis method. Leveraging the strong prior capability of a pretrained Latent Diffusion Model (LDM), GAA can synthesize realistic, diverse, and spatially aligned anomaly image–mask pairs on large-scale normal data with only a few anomaly samples.
To enhance the semantic consistency of anomaly generation, we propose a \textbf{conditional embedding mechanism focused on anomaly concept learning}. Specifically, we first introduce the \textbf{Adaptive Multi-Round Anomaly Clustering} strategy, which performs multi-round feature clustering on coarse anomaly labels to refine them into semantically purer and structurally more consistent sub-categories. This provides a clearer semantic basis for subsequent anomaly modeling. Building on this, we design the \textbf{Localized Concept Decomposition} strategy, which decomposes the anomaly concept embedding into two complementary components: a \textit{feature-focused embedding} that captures the semantic attributes of local anomalies, and a \textit{position-focused embedding} that encodes the spatial location of the anomaly. These two embeddings are jointly used as conditional inputs to guide a pretrained LDM in generating realistic and spatially aligned anomaly image–mask pairs.

%
Moreover, we propose a \textbf{region-guided anomaly mask synthesis} method that revisits and refines the mask generation process from three perspectives: structural diversity, semantic alignment, and type adaptability, enabling the creation of high-quality, spatially aligned anomaly masks. 
%
To further enhance the quality of synthesized samples, we propose a \textbf{low-quality sample filtering} method based on a new metric, ARS, which measures anomaly saliency and mask alignment. It enables automatic removal of low-quality image–mask pairs, enhancing the dataset’s applicability to real-world industrial scenarios.

Extensive experiments demonstrate that GAA achieves superior performance in realistic anomaly generation, outperforming state-of-the-art methods on both qualitative and quantitative evaluations across the MVTec AD and MVTec LOCO benchmark datasets. Leveraging high-quality synthetic data, GAA also yields significant improvements in downstream tasks such as anomaly localization and classification, effectively mitigating the challenges posed by anomaly data scarcity.
\section{Related Work}
\label{sec:related_work}
\begin{figure*}[!t]
\centering
\includegraphics[width=1.0\textwidth]{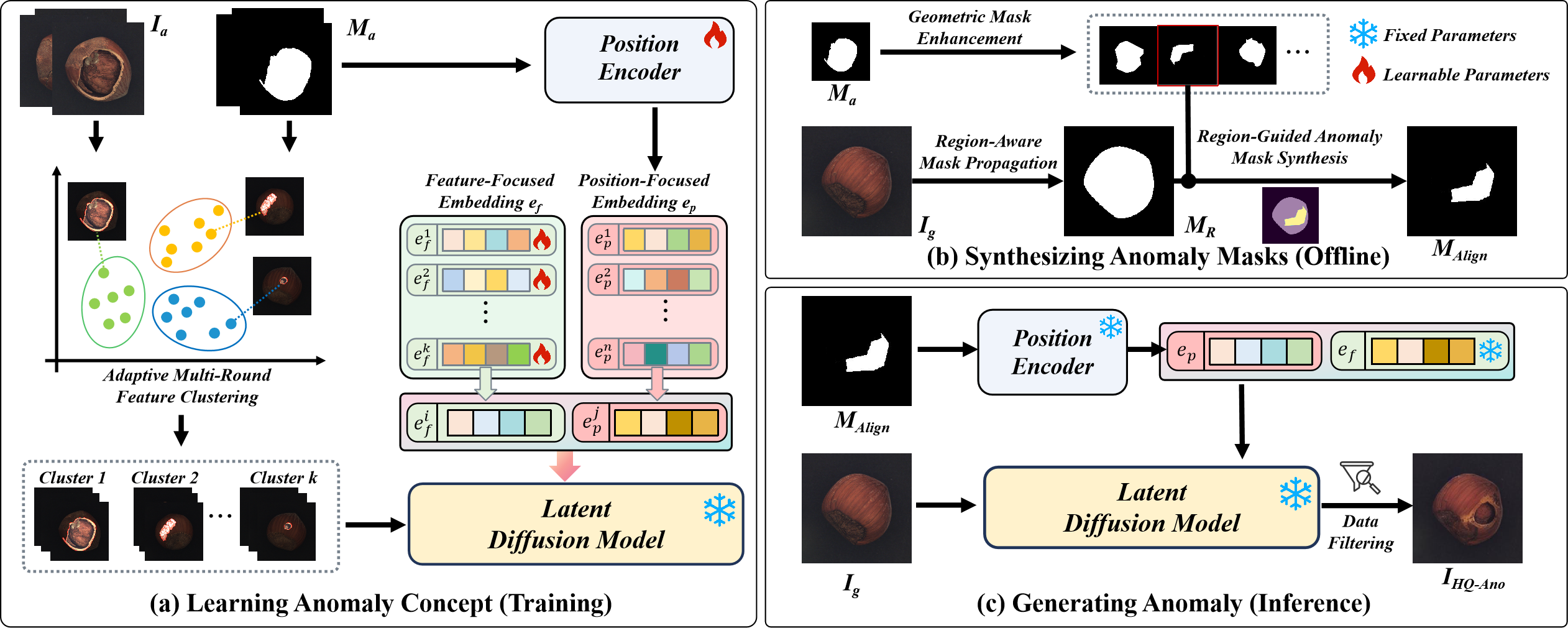}
\vspace{-15pt}
\caption{\textbf{Overview of the GAA framework.} (a) Learning anomaly concept: By training with feature-focused embeddings $e_f$ (representing anomaly appearance) and position-focused embeddings $e_p$ (representing spatial location) as textual conditions, the model learns diverse anomaly concepts across $k$ clusters. (b) Synthesizing anomaly masks: Through three refinement strategies, the model progressively generates anomaly masks $M_{Align}$ that are spatially aligned with normal images, serving as guidance for subsequent anomaly generation. (c) Generating anomaly: The learned $e_f$ and the $e_p$ encoded from $M_{Align}$ jointly guide the generation of anomalous images.}
\label{fig:framework}
\end{figure*}
\subsection{Anomaly Inspection}
The anomaly inspection task encompasses the detection, localization, and classification of anomalies. 
Given the scarcity of anomaly samples, most methods utilize unsupervised learning, training models solely with normal images.
One family of methods is based on image reconstruction, employing a variety of techniques, including autoencoders~\cite{chaudhry2018riemannian}, generative adversarial networks (GANs)~\cite{duan2023few} and diffusion models~\cite{zhang2023unsupervised, lu2023removing}. These methods identify anomalies by comparing the differences between the reconstructed images and the anomalous images. 
Another set of methods is based on feature embedding, leveraging memory networks~\cite{cohen2020sub} and normalizing flows~\cite{rudolph2021same}. These methods compress normal features into a compact space, thereby distancing anomalous features from the normal clusters within the embedding space.
While these methods excel in anomaly detection, they often struggle with anomaly localization and classification due to their unsupervised nature. Conversely, supervised anomaly detection methods~\cite{venkataramanan2020attention, pang2021explainable} are capable of classifying anomalies but are constrained by the limited availability of samples. Our method addresses these limitations by synthesizing a large number of samples, significantly enhancing the performance of supervised detection methods in downstream tasks.
\subsection{Few-Shot Personalized Generation}
In recent years, large-scale diffusion models~\cite{rombach2022high, saharia2022photorealistic} have excelled in text-to-image synthesis tasks, leveraging their powerful architectures~\cite{song2019generative, qu2025drag, li2025synergyamodal, yue2024adaptive, li2024director3d, ho2020denoising} to generate almost any image from a text prompt. However, generating personalized objects that are not prevalent in the training dataset remains a challenge, necessitating an inversion process where a specific input image is regenerated using a text-guided diffusion model.
Inversion has been extensively studied in Generative Adversarial Networks (GANs)~\cite{bermano2022state, xia2022gan}, with methods ranging from latent space optimization~\cite{abdal2019image2stylegan} and encoders~\cite{tov2021designing} to feature space encoders~\cite{wang2022high} and model fine-tuning~\cite{roich2022pivotal, alaluf2022hyperstyle}. 
For diffusion models, text inversion techniques~\cite{dhariwal2021diffusion} have shown that the DDIM sampling process~\cite{song2020denoising} can be inverted in a closed form, enabling the extraction of a latent noise map that reproduces a given real image.
However, while textual inversion~\cite{gal2022image} methods encode 3-5 images into the textual space of a pretrained latent diffusion model (LDM)~\cite{rombach2022high} for high-quality conceptual generation, they often struggle to accurately maintain unaltered portions when editing real images.
\subsection{Anomaly Synthesis}
Anomaly synthesis is a crucial technique for enhancing the performance of anomaly inspection models, particularly when actual defect data is scarce. Traditional methods, such as DRAEM~\cite{zavrtanik2021draem}, CutPaste~\cite{li2021cutpaste}, and PRN~\cite{zhang2023prototypical}, generate anomalies by cropping and pasting unrelated textures or pre-existing defects onto normal samples. However, these synthesized anomalies often lack realism and diversity. GAN-based models, such as SDGAN~\cite{niu2020defect}, Defect-GAN~\cite{zhang2021defect}, and DFMGAN~\cite{duan2023few}, improve realism by learning from defect data but require extensive anomaly datasets and face challenges in accurately generating and aligning corresponding anomaly masks. Recently, diffusion-based generation methods have demonstrated significant potential. AnoDiff~\cite{hu2024anomalydiffusion} and DualAnoDiff~\cite{jin2024dualanodiff}, leveraging pre-trained Latent Diffusion Models (LDMs), can effectively generate anomalies and corresponding masks on normal samples, even with limited training data. However, it lacks control over the generation of diverse types of anomalies and their aligned masks, limiting its generalization capability. In contrast, our proposed GAA paradigm enables controlled synthesis of high-quality anomaly-mask pairs across various anomaly types, demonstrating strong practical applicability.
\section{Method}
\label{sec:method}
We propose a novel method named \textbf{GAA} for generating diverse and realistic anomalous images along with their corresponding masks. The overall architecture is illustrated in Figure~\ref{fig:framework}. During the training phase (a), GAA leverages a pretrained Latent Diffusion Model and requires only a small number of reference anomaly image–mask pairs to effectively learn anomaly concepts. In the inference phase (c), GAA injects the learned anomaly concepts into specific regions of normal images using synthetic masks generated in (b), thereby synthesizing high-quality anomalous samples. This process significantly enhances the performance of downstream industrial anomaly detection tasks.

This section is structured as follows: Section ~\ref{sec:3.1} provides a review of the background on Latent Diffusion Model and Textual Inversion. Section ~\ref{sec:3.2} presents a detailed formulation of how GAA effectively learns anomalous concepts. Section ~\ref{sec:3.3} describes the method for synthesizing masks for different types of anomalies. Finally, Section~\ref{sec:3.4} elaborates on the process of guiding the generation of high-quality anomalous images by integrating learned anomaly concepts with the synthesized masks.
\subsection{Preliminaries}
\label{sec:3.1}
\textbf{Latent Diffusion Model (LDM)} is a text-guided diffusion model. It uses a pre-trained Variational Autoencoder (VAE)~\cite{kingma2013auto} encoder \( E \) to map an input image \( x \) into a latent representation \( z = E(x) \), where the iterative denoising process is performed in this latent space. The pre-trained VAE decoder \( D \) maps the denoised \( z \) back to pixel space to generate images.
In LDM, the conditioning input (e.g., text prompt $\mathcal{P}$) is encoded into a vector by the CLIP~\cite{radford2021learning} text encoder \( \tau_\theta \), guiding denoising. At a random timestep \( t \), noise \( \epsilon \sim N(0, I) \) is added to the latent \( z \), giving \( \hat{z}_t = \alpha_t z + \beta_t \epsilon \). The U-Net~\cite{ronneberger2015u} denoising network \( \epsilon_\theta \) removes noise iteratively to recover a clean latent. The training minimizes denoising error with the loss:
\begin{equation}
L_{\text{LDM}} = \mathbb{E}_{E(x), \epsilon, t} \| \epsilon - \epsilon_\theta(\hat{z}_t, \tau_\theta(\mathcal{P}), t) \|_2^2 ,
\end{equation}
where \( \epsilon_\theta(.,.,.) \) denotes the noise predicted by the U-Net, and \( \tau_\theta(.) \) is the conditioning vector encoded by CLIP. The loss function is optimized jointly over \( \epsilon_\theta \) and \( \tau_\theta \).
\noindent\textbf{Textual Inversion} leverages a pre-trained LDM by optimizing text embeddings to capture the characteristics of input samples. These optimized embeddings are used as conditions to extract shared content information from a few-shot input set, guiding the model to generate images consistent with the input samples.

\subsection{Learning Anomaly Concept}
\label{sec:3.2}
To effectively learn anomaly concepts from a limited set of anomaly samples, we draw inspiration from Textual Inversion and AnoDiff, and define a placeholder token $\mathbf{A}^*$ to represent the novel anomaly concept to be learned. During the embedding process, the vector associated with this token is replaced by a learnable embedding $\mathbf{A}^*$, allowing the anomaly concept to be flexibly integrated into text prompts, similar to ordinary words in natural language.
\noindent\textbf{Localized Concept Decomposition.}~Unlike~\cite{gal2022image}, which captures global semantics of an image, our method focuses on learning localized anomaly concepts within image $I_a$, guided by the ground-truth anomaly mask $M_a$. To achieve this, we decompose the embedding $\mathbf{A}^*$ into two complementary components: a feature-focused embedding $\mathbf{e}_f$, which models semantic characteristics such as texture and color of the anomaly region, and a position-focused embedding $\mathbf{e}_p$, which captures the spatial location of the anomaly in the image.
The final embedding is constructed as $\mathbf{A}^* = \{\mathbf{e}_f , \mathbf{e}_p\}$, which is then used to create conditional prompts (e.g., ``a photo of $\mathbf{A}^*$'') and fed into a pre-trained LDM to guide the generation of anomaly images. This design grants our proposed GAA framework fine-grained controllability: $\mathbf{e}_f$ controls \textit{what} type of anomaly is generated, while $\mathbf{e}_p$ controls \textit{where} the anomaly appears.

\noindent\textbf{Adaptive Multi-Round Feature Clustering.}  
Industrial anomaly datasets often lack fine-grained annotations, where semantically different anomaly types are grouped under a single label, which hinders accurate feature modeling. For example, in MVTec AD~\cite{bergmann2019mvtec}, the ``toothbrush'' category includes various types of anomalies such as \textit{``bristle deformation''}, \textit{``missing bristles''}, and \textit{``contamination''}, all labeled as \textit{``defect''}. Such label ambiguity can lead to entangled semantic embeddings.

To address this issue, we propose an adaptive multi-round clustering approach that performs fine-grained partitioning of anomaly features, thereby improving the clarity and consistency of anomaly modeling. Specifically, given an anomaly image \(I_a\) and its corresponding mask \(M_a\), we first use a pretrained ViT-B/16~\cite{dosovitskiy2020image} to extract multi-layer features. To enhance the model’s sensitivity to low-level visual cues, we additionally compute LAB~\cite{schanda2007cie} color features and LBP~\cite{ojala2002multiresolution} texture features. These features are then adaptively weighted and concatenated to form a fused anomaly feature representation \(F_a\). We then apply a multi-round KMeans clustering scheme on the coarse-grained anomaly features \(F_a\) to perform fine-grained category refinement. To improve clustering accuracy and robustness, we incorporate the Calinski-Harabasz index~\cite{calinski1974dendrite} and the Davies-Bouldin index~\cite{davies1979cluster}, which jointly consider intra-cluster compactness and inter-cluster separability. The optimal number of clusters \(K\) is determined by:
\begin{equation}
K = \arg\min_{k \geq 2} \left( 
w_1 \cdot \frac{\text{Tr}(B_k) \cdot (N - k)}{\text{Tr}(W_k) \cdot (k - 1)} + 
w_2 \cdot \frac{1}{k} \sum_{i=1}^{k} \max_{j \neq i} \left( \frac{\sigma_i + \sigma_j}{d_{ij}} \right) 
\right),
\end{equation}

where \(\text{Tr}(W_k)\) and \(\text{Tr}(B_k)\) denote the trace of the within-cluster and between-cluster scatter matrices, respectively; \(\sigma_i\) is the average distance from samples to the center of cluster \(i\); \(d_{ij}\) is the Euclidean distance between cluster centers \(i\) and \(j\); and \(N\) is the total number of samples. If the number of samples in any cluster exceeds \(\frac{N}{\log(0.1N)+0.4}\), it is regarded as a dense cluster and will be further split. This process enables hierarchical and adaptive refinement of anomaly categories.\\
\textbf{Training Pipeline.} 
For each clustered anomaly group $i \in K$, we utilize the feature-focused embedding \(e_f^i\) to represent its anomaly features, while a shared position encoder \(E\) processes the mask information for all types of anomalies. During training, a set of image-mask pairs \(\{I_a, M_a\}\) are input, and the position-focused embedding \(e_p = E(M_a)\) is extracted from the mask \(M_a\) using the position encoder \(E\). Subsequently, the feature-focused embedding \(e_f^i\) and the position embedding \(e_p^j\) are concatenated into a text embedding \(e\) = \{$e_f^i,e_p^j$\}, which serves as the conditional input for the diffusion model. The loss function and the complete training process are formulated as:
\begin{align}
\mathcal{L} &= \mathbb{E}_{\mathcal{E}(I_a),M_a , \epsilon ,t } \left\| (\epsilon - \epsilon_{\theta}(\hat{z}_t,\{e_f,e_p\}, t)  \odot M_a )\right\|_2^2 ,
\end{align}
\begin{align}
e_f^*, E^* = \arg\min_{e_f, E} \mathbb{E}_{z \sim \mathcal{E}(I_a), M_a, \epsilon, t} \mathcal{L} ,
\end{align}
Where, \(\mathcal{E}(\cdot)\) is the image encoder of the latent diffusion model, \(\hat{z}_t\) is the latent variable at time step \(t\), and \(\epsilon \sim \mathcal{N}(0,1)\) is the Gaussian noise.
\subsection{Synthesizing Anomaly Masks}
\label{sec:3.3}
Existing anomaly synthesis methods, such as random cropping~\cite{li2021cutpaste} and uncertainty-based generation strategies~\cite{duan2023few,hu2024anomalydiffusion}, often lack precise control over the anomaly masks. This misalignment between generated masks and actual anomaly regions limits their effectiveness. A straightforward and effective approach is to utilize SAM~\cite{Kirillov_2023_ICCV} to first segment the target region in a normal image, and then relocate the ground-truth (GT) anomaly masks from the dataset to this region to generate new masks. However, this approach suffers from two major limitations: 
(1)~Limited diversity: The generated anomaly masks are structurally monotonous and fail to cover the wide variability of anomalies observed in industrial settings;  
(2)~Poor scalability: The SAM-based interactive selection of target regions is only applicable to a small number of samples, leading to high labor costs when scaling to large normal datasets.
To address the above challenges, we propose a region-guided mask synthesis method that improves the mask generation process from three perspectives: structural diversity enhancement, semantic region alignment, and anomaly type adaptation. This method consists of the following three strategies:

\noindent\textbf{Geometric Mask Enhancement.}  
This module enhances the structural diversity of anomaly masks via morphological transformations. Specifically, we first apply a closing operation (i.e., dilation $M_a \oplus B_1$ followed by erosion $\ominus B_2$) to the original anomaly mask $M_a$ to fill small holes and smooth the boundaries. Then, we extract its contours $\mathcal{F}(\cdot)$ and employ polygon approximation $\mathcal{P}(\cdot, \delta)$ along with random perturbations $\Delta P$ to construct a structurally diverse pseudo-mask $M_{\text{opt}}$, defined as:
\begin{equation}
M_{\text{opt}} = \mathcal{P} \left( \mathcal{F} \left( \left( M_a \oplus B_1 \right) \ominus B_2 \right), \delta \right) + \Delta P,
\end{equation}
Furthermore, to ensure consistency in area with the average anomaly area $A_{\text{avg}}$ of the corresponding category, we perform scale-adaptive adjustment on $M_{\text{opt}}$ to obtain the final optimized mask $M_G$, defined as:
\begin{equation}
M_G^{(x, y)} = M_{\text{opt}}^{(x, y)} \cdot \left( \alpha \cdot \sqrt{\frac{A_{\text{avg}}}{A_M}} + (1 - \alpha) \cdot \sqrt[3]{\frac{A_{\text{avg}}}{A_M}} \right),
\end{equation}
\noindent where $A_M$ denotes the area of the current mask, $\alpha$ is a scaling factor, and $(x, y)$ represents the pixel coordinates.

\noindent\textbf{Region-Aware Mask Propagation.}  
To achieve precise alignment of anomaly masks with semantic regions, we propose a region-guided mask generation module. Based on the distribution of anomalies across different product categories, we predefine a set of semantic regions $\{R_j\}_{j=1}^{J}$ (e.g., a "breakfast box" might be divided into regions such as nuts, cereal, and container).

Next, for each product category, we randomly select 3$\sim$5 normal samples $I_g^{\text{Few}}$ and obtain initial region masks $M_{init}$ using SAM~\cite{Kirillov_2023_ICCV}. These masks are used as prompts for a pretrained prompt-based segmentation model, SegGPT~\cite{SegGPT}, to infer semantic region masks $M_R$ across all normal samples of that category, thereby enabling automated propagation and alignment of regional information.

\begin{figure}[!t]
\centering
\includegraphics[width=0.48\textwidth]{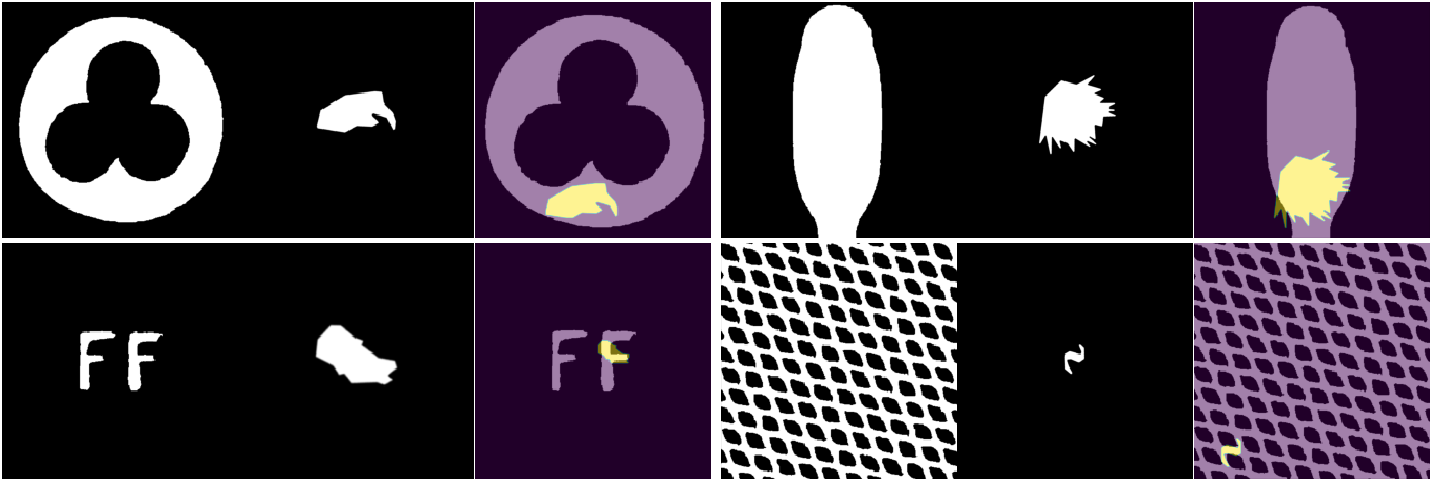}
\vspace{-10pt}
 \caption{Illustration of region-guided anomaly mask synthesis. The purple areas indicate the semantic regions used for guidance $M_{\text{R}}$, while the yellow areas represent the aligned anomaly masks $M_{\text{Align}}$.}
\vspace{-15pt}
\label{fig:heatmap}
\end{figure}
\noindent\textbf{Region-Guided Anomaly Mask Synthesis.}  
Having obtained the optimized anomaly mask $M_G$ and the semantic region mask $M_R$, we further propose a region-guided mask synthesis strategy to generate aligned anomaly masks $M_{\text{Align}}$. Given the complexity of industrial anomalies, naive random placement fails to capture the diversity of real-world scenarios. Referring to the anomaly type taxonomy in the MVTec dataset~\cite{bergmann2022beyond,hu2024anomalyxfusion}, we divide anomalies into three categories and design corresponding synthesis strategies:~(1)\textit{Structural Anomalies}: These include physical anomalies such as cracks or scratches. We synthesize $M_{\text{Struct}}$ by embedding $M_G$ at random locations or along the edges of $M_R$.~(2)\textit{Logical Anomalies}:~Examples include missing or misplaced components. Here, we directly use the predefined semantic region mask $M_R$ as the logical anomaly mask $M_{\text{Logic}}$, e.g., mapping “missing nuts” to the designated nut region.~(3)\textit{Combined Anomalies}:
A single product may simultaneously contain multiple types of anomalies. We randomly combine multiple masks, ensuring spatial non-overlap, to generate the combined anomaly mask $M_{\text{Cob}}$. Finally, we obtain a collection of structurally sound and semantically aligned anomaly masks over the entire set of normal samples, denoted as:  
\(M_{\text{Align}} = \{ M_{\text{Struct}}, M_{\text{Logic}}, M_{\text{Cob}} \}.\)
\subsection{Generating Anomaly}
\label{sec:3.4}
\noindent\textbf{Inference Pipeline.}  
During the anomaly concept modeling stage, the model learns semantic representations corresponding to various anomaly feature clusters, and captures their characteristics through a feature-focused embedding $e_f$. In the anomaly mask generation stage, semantically aligned anomaly masks $M_\text{Align}$ are synthesized for each normal sample and transformed into position-focused embeddings $e_p$ via a trained position encoder $E$, enabling precise control over the spatial distribution of anomalies.
Finally, the feature embedding $e_f$ and position embedding $e_p$ are concatenated into a unified control condition $e = \{e_f, e_p\}$, which is fed into the latent diffusion model to enable high-fidelity, controllable, and large-scale synthesis of diverse anomaly types on normal images $I$.

\noindent\textbf{Low-Quality Sample Filtering.}  
Due to the inherent uncertainty of generative models, it is inevitable that some low-quality samples are produced during the synthesis process. Inspired by unsupervised anomaly detection methods~\cite{liu2023simplenet}, we propose an automated anomaly filtering strategy.

Specifically, we use a feature extractor $F_\beta$ to encode normal images and construct pseudo-anomaly features by adding Gaussian noise. These are used to train a simple binary anomaly discriminator $B_D$.
Subsequently, features are extracted from the generated anomaly image-mask pairs $\{I_g, M_\text{Align}\}$ and passed into $B_D$ to obtain an anomaly score map $S$. Conventional unsupervised methods often assess anomaly saliency using the maximum score, which fails to capture the overall quality of the anomaly region. To address this, we propose a new metric called Anomaly Region Score (ARS), which jointly evaluates the relative saliency and spatial alignment of generated anomalies with their corresponding masks. ARS is defined as follows:

\begin{equation}
ARS = \frac{\sum_{i=1}^{H} \sum_{j=1}^{W} [U(B_D(F_\beta(I_g))) \odot M_\text{Align}](i,j)}{\sum_{i=1}^{H} \sum_{j=1}^{W} M_\text{Align}(i,j)},
\label{eq:ars}
\end{equation}

where $U$ denotes the upsampling operation, and $H$ and $W$ represent the image height and width, respectively. Based on the ARS score, we select higher-quality anomaly image–mask pairs $\{I_{\text{HQ-Ano}}, M_{\text{Align}}\}$, which significantly improve the performance of downstream tasks.

\begin{table*}[t]
    \centering
    \caption{Quantitative comparison on the MVTec AD dataset. Our model achieves the best performance in both image quality and diversity, obtaining the highest IS and IC-LPIPS.}
    \vspace{-5pt}
    \renewcommand{\arraystretch}{0.82} 
    \setlength{\tabcolsep}{5pt} 
    \resizebox{0.88\linewidth}{!}{%
    \begin{Large}
    \begin{tabular}{c|cc|cc|cc|cc|cc}
        \toprule
        \multirow{2}{*}{Category} & \multicolumn{2}{c|}{Defect-GAN} & \multicolumn{2}{c|}{Cut-Paste} & \multicolumn{2}{c|}{DFMGAN} & \multicolumn{2}{c|}{AnoDiff} & \multicolumn{2}{c}{Ours} \\
        \cmidrule(lr){2-11}
       & IS~$\uparrow$ &  IC-LPIPS~$\uparrow$ & IS~$\uparrow$ &  IC-LPIPS~$\uparrow$ & IS~$\uparrow$ &  IC-LPIPS~$\uparrow$ & IS~$\uparrow$ &  IC-LPIPS~$\uparrow$ & IS~$\uparrow$ &  IC-LPIPS~$\uparrow$ \\
        \midrule
bottle & 1.39 & 0.07 & 1.48 & 0.05 & 1.62 & 0.12 & 1.58 & 0.19 & \textbf{2.08} & \textbf{0.28} \\
cable & 1.70 & 0.22 & 1.82 & 0.26 & 1.96 & 0.25 & 2.13 & 0.41 & \textbf{2.36} & \textbf{0.47} \\
capsule & \textbf{1.59} & 0.04 & 1.46 & 0.06 & \textbf{1.59} & 0.11 & \textbf{1.59} & 0.21 & 1.53 & \textbf{0.26} \\
carpet & 1.24 & 0.12 & 1.13 & 0.11 & \textbf{1.23} & 0.13 & 1.16 & \textbf{0.24} & 1.22 & \textbf{0.24} \\
grid & 2.01 & 0.12 & \textbf{2.16} & 0.13 & 1.97 & 0.13 & 2.04 & 0.44 & 2.15 & \textbf{0.47} \\
hazel nut & 1.87 & 0.19 & 1.90 & 0.22 & 1.93 & 0.24 & 2.13 & \textbf{0.31} & \textbf{2.35} & 0.30 \\
leather & \textbf{2.12} & 0.14 & 1.72 & 0.15 & 2.06 & 0.17 & 1.94 & 0.41 & 1.91 & \textbf{0.43} \\
metal nut & 1.47 & 0.30 & \textbf{1.70} & 0.16 & 1.49 & \textbf{0.32} & 1.96 & 0.30 & 1.87 & 0.29 \\
pill & \textbf{1.61} & 0.10 & \textbf{1.72} & 0.17 & 1.63 & 0.16 & \textbf{1.61} & \textbf{0.26} & 1.57 & 0.24 \\
screw & 1.19 & 0.12 & \textbf{1.30} & 0.16 & 1.12 & 0.14 & 1.28 & 0.30 & 1.24 & \textbf{0.34} \\
tile & 2.35 & 0.22 & 1.41 & 0.18 & 2.39 & 0.22 & 2.54 & 0.55 & \textbf{2.68} & \textbf{0.59} \\
toothbrush & \textbf{1.85} & 0.03 & 1.48 & 0.08 & \textbf{1.82} & 0.18 & 1.68 & 0.21 & 1.76 & \textbf{0.23} \\
transistor & 1.47 & 0.13 & 1.50 & 0.16 & \textbf{1.64} & 0.25 & 1.57 & 0.34 & 1.54 & \textbf{0.35} \\
wood & 2.19 & 0.29 & 1.70 & 0.22 & 2.12 & 0.35 & 2.33 & 0.37 & \textbf{2.43} & \textbf{0.42} \\
zipper & 1.25 & 0.10 & 1.33 & 0.12 & 1.29 & \textbf{0.27} & 1.39 & 0.25 & \textbf{1.64} & 0.35 \\

\midrule
\textbf{Average} & 1.69 & 0.15 & 1.59 & 0.15 & 1.72 & 0.20 & 1.80 & 0.32 & \textbf{1.89} & \textbf{0.35} \\
        \bottomrule
    \end{tabular}
    \end{Large}
    }

    \label{tab:is_il_ars}
    \vspace{-5pt}
\end{table*}

\begin{figure}[t]
\centering
\includegraphics[width=0.48\textwidth]{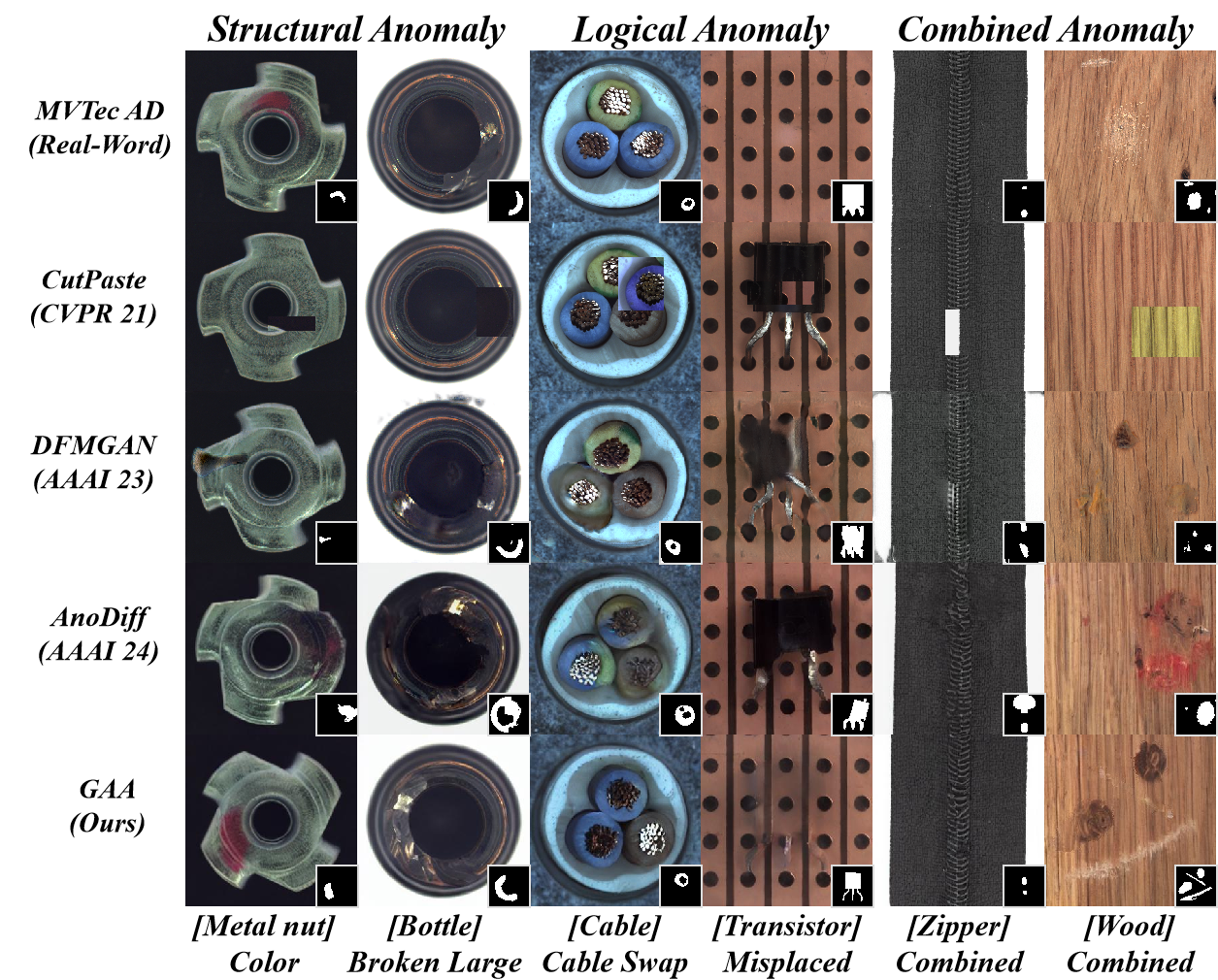}
\caption{Qualitative comparison. This figure compares our method with other approaches for generating anomaly images. Our method produces more realistic anomalies that integrate seamlessly into the objects, regardless of their structural, logical, or combined nature.}
\vspace{-15pt}
\label{fig:Comparison-of-the-generated-results}
\end{figure}

\section{Experiments}
\label{sec:exp}
\subsection{Experimental Settings}
\textbf{Dataset}.
We conducted experimental evaluations on the MVTec AD dataset~\cite{bergmann2019mvtec} and the MVTec LOCO dataset~\cite{bergmann2022beyond}.
The MVTec AD dataset comprises 15 industrial product categories and includes nearly one hundred coarsely categorized anomaly types, making it well-suited for simulating real-world industrial Anomaly scenarios. 
The MVTec LOCO dataset contains 5 product categories and features numerous structural and logical anomalies without fine-grained categorization, along with complex multi-instance cases, posing greater challenges to the generalization and robustness of anomaly detection models.\\
 \noindent\textbf{Implementation Details.} We followed the experimental setup of AnoDiff~\cite{hu2024anomalydiffusion}, using $\frac{1}{3}$ anomaly images and their corresponding masks for training. During inference, we generated 1,000 pairs of anomaly image-mask samples for each anomaly type and filtered the top 500 high-quality samples based on our proposed ARS for downstream tasks. Due to space constraints, only a subset of downstream experiments on the MVTec LOCO dataset are presented in the main text. More details are given in \textit{Appendix}.

\noindent\textbf{Evaluation Metrics.} \textbf{(1) For anomaly generation}, we measure the fidelity and diversity of generated images using Inception Score (IS)~\cite{salimans2016improved} and Intra-cluster Pairwise LPIPS Distance (IC-LPIPS)~\cite{ojha2021few}.~\textbf{ (2) For anomaly inspection}, we use AUROC, Average Precision (AP), and F1-max to assess performance on detection and localization tasks, and accuracy (ACC) to evaluate classification precision.
\begin{table*}[t]
\centering
\renewcommand{\arraystretch}{0.90}
\setlength\tabcolsep{2.0pt}
\caption{Localization Comparison. The table presents AUROC, AP, F1-max, and PRO scores for localization evaluation using a UNet trained on generated anomaly images. Our method achieves the best performance across all metrics on both the MVTec AD and MVTec LOCO datasets.}
\resizebox{1.0\linewidth}{!}{
\begin{tabular}{c|c|cccc|cccc|cccc|cccc}
\toprule
\multirow{2}{*}{Dataset} & \multirow{2}{*}{Category}  & \multicolumn{4}{c|}{DRAEM} & \multicolumn{4}{c|}{DFMGAN} & \multicolumn{4}{c|}{AnoDiff} & \multicolumn{4}{c}{Ours} \\
 \cmidrule(r){3-6} \cmidrule(r){7-10} \cmidrule(r){11-14} \cmidrule(r){15-18}
 &  & AUROC & AP & $F_1$-max & PRO & AUROC & AP & $F_1$-max & PRO & AUROC & AP & $F_1$-max & PRO & AUROC & AP & $F_1$-max & PRO \\
\midrule
\multirow{9}{*}{\shortstack{MvTec AD \\ (Objects)}} & bottle & 96.7 & 80.2 & 74.0 & 91.2 & 98.9 & 90.2 & 83.9 & 91.7 & 99.4 & 94.1 & 87.3 & 94.3 & \textbf{99.7} & \textbf{96.2} & \textbf{88.5} & \textbf{95.4} \\
 & cable & 80.3 & 21.8 & 28.3 & 58.2 & 97.2 & 81.0 & 75.4 & 84.9 & 99.2 & 90.8 & 83.5 & 95.0 & \textbf{99.6} & \textbf{93.0} & \textbf{85.8} & \textbf{96.3} \\
 & capsule & 76.2 & 25.5 & 32.1 & 81.1 & 79.2 & 26.0 & 35.0 & 66.1 & 98.8 & 57.2 & 59.8 & 95.4 & \textbf{99.0} & \textbf{68.5} & \textbf{66.8} & \textbf{96.0} \\
 & hazelnut & 98.8 & 73.6 & 68.5 & 95.9 & 99.7 & 95.2 & 89.5 & 96.4 & \textbf{99.8} & 96.5 & 90.6 & 97.1 & \textbf{99.8} & \textbf{97.5} & \textbf{91.8} & \textbf{98.2} \\
 & metal nut & 96.9 & 84.2 & 74.5 & 90.4 & 99.3 & 98.1 & 94.5 & 88.0 & \textbf{99.8} & \textbf{98.7} & 94.0 & 94.8 & \textbf{99.8} & 98.4 & \textbf{94.1} & \textbf{97.6} \\
 & pill & 95.8 & 45.3 & 53.0 & 83.7 & 81.2 & 67.8 & 72.6 & 56.5 & \textbf{99.8} & 97.0 & 90.8 & 97.3 & \textbf{99.8} & \textbf{98.5} & \textbf{90.2} & \textbf{97.9} \\
 & screw & 91.0 & 30.1 & 35.7 & 78.1 & 58.8 & 2.2 & 5.3 & 41.8 & 97.0 & 51.8 & 50.9 & 80.3 & \textbf{97.8} & \textbf{57.5} & \textbf{56.1} & \textbf{85.2} \\
 & toothbrush & 93.8 & 29.5 & 28.4 & 75.1 & 96.4 & 75.9 & 72.6 & 74.3 & 99.2 & 76.5 & 73.4 & 91.4 & \textbf{99.3} & \textbf{78.5} & \textbf{74.3} & \textbf{92.9} \\
 & transistor & 76.5 & 31.7 & 24.2 & 54.3 & 96.2 & 81.2 & 77.0 & 65.5 & \textbf{99.3} & \textbf{92.6} & 85.7 & 96.2 & 99.0 & 91.8 & \textbf{86.2} & \textbf{97.1} \\
\midrule
\multirow{6}{*}{\shortstack{MvTec AD \\ (Textures)}} & carpet & 92.6 & 43.0 & 41.9 & 80.0 & 90.6 & 33.4 & 38.1 & 76.5 & 98.6 & 81.2 & 74.6 & 91.6 & \textbf{99.3} & \textbf{86.3} & \textbf{77.4} & \textbf{92.7} \\
 & grid & 99.1 & 59.3 & 58.7 & 95.8 & 75.2 & 14.3 & 20.5 & 52.3 & 98.3 & 52.9 & 54.6 & 92.3 & \textbf{99.3} & \textbf{58.5} & \textbf{60.5} & \textbf{93.9} \\
 & leather & 98.5 & 67.6 & 65.0 & 96.7 & 98.5 & 68.7 & 66.7 & 96.0 & \textbf{99.8} & 79.6 & 71.0 & 98.2 & \textbf{99.8} & \textbf{84.3} & \textbf{74.9} & \textbf{99.1} \\
 & tile & 98.5 & 93.2 & 87.8 & 95.3 & 99.5 & 97.1 & 91.6 & 97.5 & 99.2 & 93.9 & 86.2 & 96.1 & \textbf{99.8} & \textbf{97.2} & \textbf{89.6} & \textbf{97.5} \\
 & wood & 98.8 & 87.8 & 80.9 & 94.7 & 95.3 & 70.7 & 65.8 & 89.9 & 98.9 & 84.6 & 74.5 & 94.3 & \textbf{99.2} & \textbf{88.9} & \textbf{80.6} & \textbf{96.3} \\
 & zipper & 93.4 & 65.4 & 64.7 & 84.6 & 92.9 & 65.6 & 64.9 & 83.0 & 99.4 & 86.0 & 79.2 & 96.3 & \textbf{99.8} & \textbf{87.8} & \textbf{80.9} & \textbf{96.9} \\
\midrule
\multirow{5}{*}{MvTec LOCO} & breakfast box & 89.3 & 64.5 & 61.2 & 62.5 & 92.1 & 66.8 & 65.4 & 68.5 & 92.7 & 73.0 & 69.3 & 70.2 & \textbf{95.6} & \textbf{81.4} & \textbf{77.3} & \textbf{76.4} \\
 & juice bottle & 88.5 & 48.4 & 49.3 & 80.1 & 86.9 & 40.9 & 48.3 & 78.8 & 89.2 & 51.0 & 47.7 & 82.2 & \textbf{93.5} & \textbf{57.8} & \textbf{53.5} & \textbf{85.8} \\
 & pushpins & 80.5 & 28.4 & 31.0 & 54.3 & 68.1 & 4.7 & 11.0 & 32.4 & 77.7 & 9.6 & 17.5 & 45.7 & \textbf{91.7} & \textbf{44.6} & \textbf{49.5} & \textbf{66.8} \\
 & screw bag & 70.4 & 18.5 & 24.6 & 57.4 & 63.8 & 11.4 & 20.7 & 53.1 & 68.6 & 15.0 & 24.7 & 58.7 & \textbf{74.9} & \textbf{30.6} & \textbf{34.8} & \textbf{68.9} \\
 & connectors & 71.3 & 34.5 & 35.9 & 67.4 & 62.0 & 20.4 & 25.9 & 60.3 & 67.2 & 28.3 & 29.8 & 64.8 & \textbf{78.6} & \textbf{37.5} & \textbf{38.9} & \textbf{70.4} \\
\midrule
MvTec & Average & 88.9 & 51.6 & 51.0 & 78.8 & 86.6 & 55.6 & 56.2 & 72.7 & 94.1 & 70.5 & 67.3 & 86.6 & \textbf{96.3} & \textbf{76.8} & \textbf{72.6} & \textbf{90.1} \\
\bottomrule
\end{tabular}
}
\vspace{-10pt}
\label{tab:main_exp}
\end{table*}

\begin{table}[t]
    \centering
    \renewcommand{\arraystretch}{0.8} 
    \setlength{\tabcolsep}{4pt} 
    \caption{Classification Comparison. This table presents the classification accuracy (\%) for anomaly category prediction.
AnoDiff*: results reported in the original paper; AnoDiff‡: scores re-evaluated from the officially released dataset (with low-quality samples filtered).}
    \resizebox{\linewidth}{!}{%
    \begin{tabular}{c|ccccc}
        \toprule
        Category & DefGAN & DFMGAN & AnoDiff* & AnoDiff‡ & Ours \\
        \midrule
        bottle     & 53.5 & 56.6 & 90.7 & 92.1 & \textbf{95.4} \\
        cable      & 21.4 & 45.3 & 67.2 & 78.5 & \textbf{92.2} \\
        capsule    & 32.0 & 37.2 & 66.7 & 60.1 & \textbf{69.3} \\
        carpet     & 29.0 & 47.3 & 58.1 & 65.7 & \textbf{76.0} \\
        grid       & 27.5 & 40.8 & 42.5 & 53.4 & \textbf{78.5} \\
        hazelnut   & 61.1 & 81.9 & 85.4 & 88.9 & \textbf{93.8} \\
        leather    & 42.3 & 49.7 & 61.9 & 68.3 & \textbf{85.8} \\
        metal nut  & 56.8 & 64.6 & 59.4 & 76.2 & \textbf{96.9} \\
        pill       & 28.5 & 29.5 & 59.4 & 62.1 & \textbf{69.8} \\
        screw      & 28.8 & 37.5 & 48.2 & 56.7 & \textbf{73.0} \\
        tile       & 26.9 & 74.9 & 84.2 & 89.4 & \textbf{96.7} \\
        transistor & 35.7 & 52.4 & 60.7 & 67.8 & \textbf{92.9} \\
        wood       & 24.6 & 49.2 & 71.4 & 77.0 & \textbf{88.1} \\
        zipper     & 18.7 & 27.6 & \textbf{69.5} & 64.3 & 76.9 \\
        \midrule
        Average    & 34.8 & 49.6 & 66.1 & 71.5 & \textbf{84.7} \\
        \bottomrule
    \end{tabular}
    }
    \label{tab:classification_results}
\end{table}

\subsection{Evaluation in Anomaly Generation}
\textbf{Quantitative Evaluation.} To validate the effectiveness of our generative paradigm, we selected methods such as CutPaste~\cite{li2021cutpaste}, Defect-GAN~\cite{zhang2021defect}, DFMGAN~\cite{duan2023few}, and AnoDiff~\cite{hu2024anomalydiffusion} as benchmarks for comparison.
The quantitative results for anomaly generation quality and diversity on the MVTec AD dataset are presented in Tab.\ref{tab:is_il_ars}. For each anomaly category, we employed our GAA to synthesize 500 anomaly image-mask pairs for calculating IS, IC-LPIPS, and ARS. The results demonstrate that our GAA achieves the best overall performance in terms of fidelity, diversity, and quality of the synthesized anomaly image-mask pairs.

\noindent\textbf{Qualitative Evaluation.} 
We compare the anomaly images generated by different methods on the MVTec AD dataset, as shown in Figure~\ref{fig:Comparison-of-the-generated-results}. CutPaste produces unrealistic anomalies, while DFMGAN suffers from low fidelity and poor mask alignment. AnoDiff tends to generate semantically entangled anomalies—for instance, irregular black fragments appear within red contamination in the wood category. It also exhibits mask misalignment issues, such as oversized and incomplete masks in the cable category. In contrast, our GAA effectively synthesizes diverse anomaly types with stronger generalization. The generated anomalies are both more realistic and better aligned with their corresponding masks.

\noindent\textbf{Anomaly Generation For Localization.}
We conducted a comparative evaluation of several mainstream anomaly generation methods, including DRAEM~\cite{Zavrtanik_2021_ICCV}, DFMGAN~\cite{duan2023few}, and AnoDiff~\cite{hu2024anomalydiffusion}, to assess the performance of their generated anomaly image–mask pairs in anomaly localization tasks. For each anomaly type, we synthesized 500 pairs of anomaly images and masks, which were then used to train a U-Net network~\cite{ronneberger2015unet} .
As shown in Table~\ref{tab:main_exp} our GAA method outperforms existing approaches on both MVTec AD and LOCO datasets. It achieves an AUROC of \textbf{96.3\% (+2.2\%)} and an AP of \textbf{76.8\% (+6.3\%)}, demonstrating superior performance and stronger generalization in anomaly localization tasks.
In addition, Figure~\ref{fig:heatmap} shows the qualitative comparison of anomaly localization results, where our method demonstrates clearly superior performance in identifying anomaly regions.

%
\begin{table*}[t]
    \centering
    \renewcommand{\arraystretch}{0.85} 
    \setlength{\tabcolsep}{3pt} 
    \caption{Comparison of pixel-level anomaly localization (AUROC/AP) on MVTec AD between a simple U-Net trained on our generated dataset and existing anomaly detection methods using their official codes or pre-trained models.}
\vspace{-7pt}
    \resizebox{\linewidth}{!}{%
    \begin{Large}
    \begin{tabular}{@{}c|ccccccccccc@{}}
        \toprule
        \multirow{2}{*}{Category} & \multicolumn{7}{c}{Unsupervised} & \multicolumn{4}{c}{Supervised} \\
        \cmidrule(lr){2-8}\cmidrule(lr){9-12} 
       & CFLOW & DRAEM & SSPCAB & CFA & RD4AD & PatchCore & Musc & DevNet & DRA & PRN & Ours \\ 
        \midrule
        bottle & 98.8/49.9 & 99.1/88.5 & 98.9/88.6 & 98.9/50.9 & 98.8/51.0 & 97.6/75.0 & 98.5/82.8 & 96.7/67.9 & 91.7/41.5 & 99.4/92.3 & \textbf{99.7/96.2} \\
cable & 98.9/72.6 & 94.8/61.4 & 93.1/52.1 & 98.4/79.8 & 98.8/77.0 & 96.8/65.9 & 96.2/58.8 & 97.9/67.6 & 86.1/34.8 & 98.8/78.9 & \textbf{99.6/93.0} \\
capsule & \textbf{99.5}/64.0 & 97.6/47.9 & 90.4/48.7 & 98.9/71.1 & 99.0/60.5 & 98.6/46.6 & 98.9/52.7 & 91.1/46.6 & 88.5/11.0 & 98.5/62.2 & 99.0/\textbf{68.5} \\
carpet & \textbf{99.7}/67.0 & 96.3/62.5 & 92.3/49.1 & 99.1/47.7 & 99.4/46.0 & 98.7/65.0 & 99.4/75.3 & 94.6/19.6 & 98.2/54.0 & 99.0/82.0 & 99.3/\textbf{86.3} \\
grid & 99.1/\textbf{87.8} & \textbf{99.5}/53.2 & 99.6/58.2 & 98.6/82.9 & 98.0/75.4 & 97.2/23.6 & 98.6/37.0 & 90.2/44.9 & 86.2/28.6 & 98.4/45.7 & 99.3/58.5 \\
hazelnut & 97.9/67.2 & 99.5/88.1 & 99.6/94.5 & 98.5/80.2 & 94.2/57.2 & 97.6/55.2 & 99.3/74.4 & 76.9/46.8 & 88.8/20.3 & 99.7/93.8 & \textbf{99.8/97.5} \\
leather & 99.2/\textbf{91.1} & 98.8/68.5 & 97.2/60.3 & 96.2/60.9 & 96.6/53.5 & 98.9/43.4 & 99.7/62.6 & 94.3/66.2 & 97.2/5.1 & 99.7/69.7 & \textbf{99.8}/84.3 \\
metal nut & 98.8/78.2 & 98.7/91.6 & 99.3/95.1 & 98.6/74.6 & 97.3/53.8 & 97.5/86.8 & 87.5/49.6 & 93.3/57.4 & 80.3/30.6 & 99.7/98.0 & \textbf{99.8/98.4} \\
pill & 98.9/60.3 & 97.7/44.8 & 96.5/48.1 & 98.8/67.9 & 98.4/58.1 & 97.0/75.9 & 97.6/65.6 & 98.9/79.9 & 79.6/22.1 & 99.5/91.3 & \textbf{99.8/98.5} \\
screw & 98.8/45.7 & \textbf{99.7/72.9} & 99.1/62.0 & 98.7/61.4 & 99.1/51.8 & 98.7/34.2 & 98.9/31.7 & 66.5/21.1 & 51.0/5.1 & 97.5/44.9 & 97.8/57.5 \\
tile & 98.0/86.7 & 99.4/96.4 & 99.2/96.3 & 98.6/92.6 & 97.4/78.2 & 94.9/56.0 & 98.3/80.6 & 88.7/63.9 & 91.0/54.4 & 99.6/96.5 & \textbf{99.8/97.2} \\
toothbrush &  99.1/56.9 & 97.3/49.2 & 97.5/38.9 & 98.4/61.7 & 99.0/63.1 & 97.6/37.1 & \textbf{99.4}/64.2 & 96.3/52.4 & 74.5/4.8 & 99.6/78.1 & 99.3/\textbf{78.5} \\
transistor & 98.8/40.6 & 92.2/56.0 & 85.3/36.5 & 98.6/82.9 & 99.6/50.3 & 91.8/66.7 & 92.5/61.2 & 55.2/4.4 & 79.3/11.2 & 98.4/85.6 & \textbf{99.0/91.8} \\
wood & 98.9/47.2 & 97.6/81.6 & 97.2/77.1 & 97.6/25.6 & 99.3/39.1 & 95.7/54.3 & 98.7/77.5 & 93.1/47.9 & 82.9/21.0 & 97.8/82.6 & \textbf{99.2/88.9} \\
zipper & 96.5/63.9 & 98.6/73.6 & 98.1/78.2 & 95.9/53.9 & 99.7/52.7 & 98.5/63.1 & 98.4/64.2 & 92.4/53.1 & 96.8/42.3 & 98.8/77.6 & \textbf{99.8/87.8} \\
\midrule
\textbf{Average} & 98.7/65.3 & 97.7/69.0 & 96.2/65.5 & 98.3/66.3 & 98.3/57.8 & 97.1/56.6 & 97.5/62.6 & 86.4/49.3 & 84.8/25.7 & 99.0/78.6 & \textbf{99.4/85.9} \\

        \bottomrule
    \end{tabular}
    \end{Large}
    }
    \vspace{-5pt}
    \label{tab:pixel_level_anomaly}
\end{table*}

\noindent\textbf{Anomaly Generation for Classification.}
Anomaly classification evaluation reflects the distinctiveness and realism of generated samples across categories. To validate the effectiveness of our generation model, we followed the experimental setting of previous work~\cite{hu2024anomalydiffusion} and used a ResNet-34~\cite{he2016deep} classifier for performance evaluation. The classifier was trained on synthetic data and evaluated on a unified test set.
As shown in Table~\ref{tab:classification_results}, our proposed GAA method outperformed all baselines across all categories on the MVTec AD dataset, with particularly strong performance on challenging classes such as metal nuts, transistors, and cables. GAA achieved a classification accuracy of \textbf{84.7\%}, representing an \textbf{18.6\%} improvement over the state-of-the-art AnoDiff. These results strongly demonstrate the superior quality of the anomaly data generated by our paradigm.
\subsection{Comparison with Detection Models}
To assess the practical value of GAA, we compared it with several state-of-the-art anomaly detection methods~\cite{ristea2022self,deng2022anomaly,roth2022towards,ding2022catching,zhang2023prototypical}. All models were evaluated on a consistent test set containing two-thirds of the anomalies from MVTec AD, using official code or publicly available pre-trained models. For RPN, we reported the results from its original paper due to the lack of public code.

As shown in Table~\ref{tab:pixel_level_anomaly}, our method, using a simple U-Net~\cite{ronneberger2015unet}, achieved \textbf{99.4\%} AUROC and \textbf{85.9\%} AP for anomaly localization, outperforming all existing methods. These results demonstrate the significance of high-quality synthetic anomalies in industrial applications.
\begin{figure}[!t]
\centering
\includegraphics[width=0.45\textwidth]{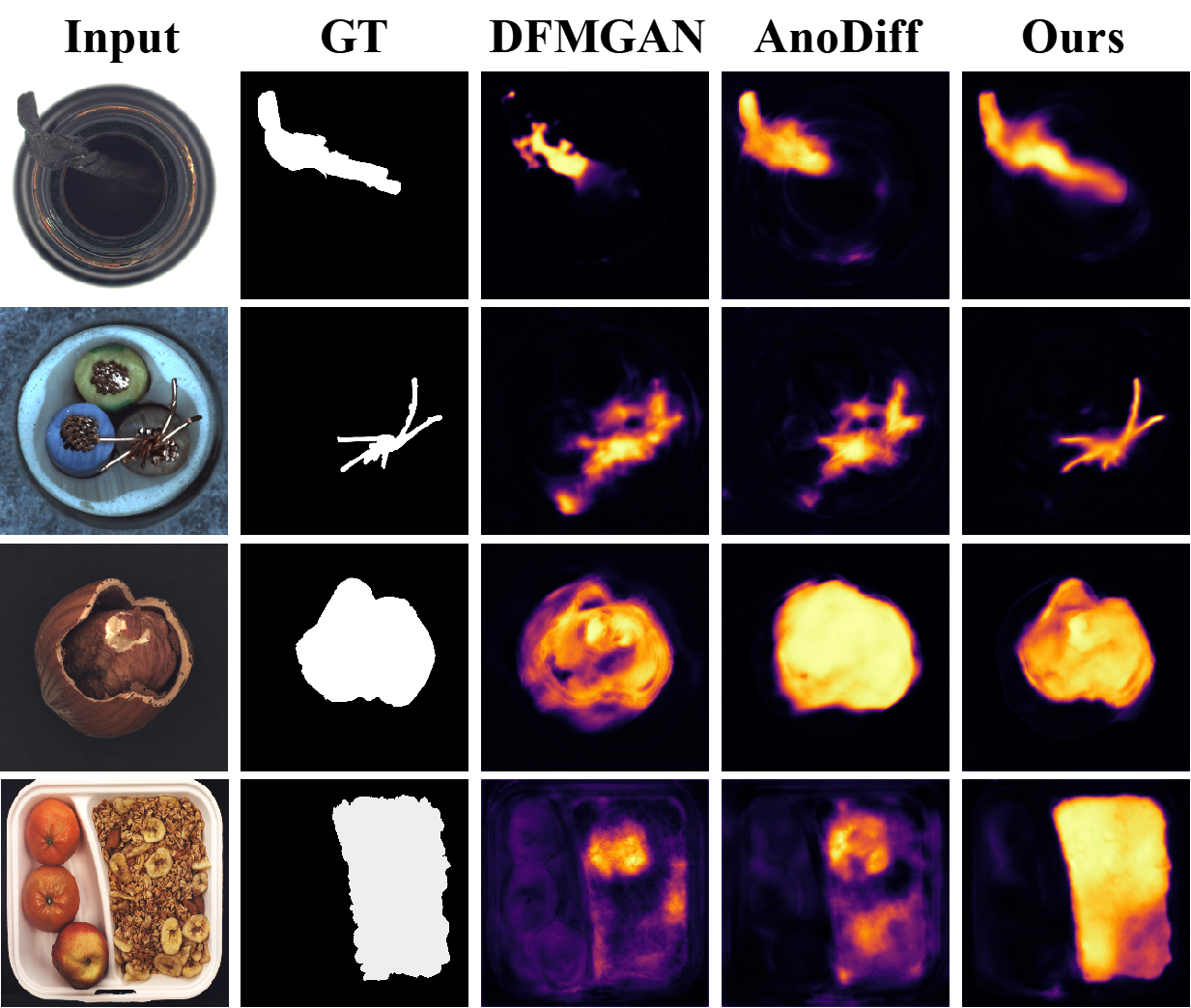}
\caption{\textbf{Quantitative comparison of anomaly localization using U-Net trained on data from DFMGAN, AnoDiff, and our model.} }
\vspace{-10pt}
\label{fig:heatmap}
\end{figure}

\subsection{Ablation Study}
\textbf{Module Ablation.} Table~\ref{tab:table_abalation} presents the impact of stepwise removal of each module on the anomaly synthesis performance of the GAA framework.~\textbf{(1) Adaptive Multi-Round Feature Clustering (AMFC):} Introducing AMFC significantly improves ACC and IS, indicating its effectiveness in enabling the model to learn more diverse and distinctive anomaly features.~\textbf{(2) Region-Guided Anomaly Mask Synthesis (RAMS):} The inclusion of RAMS leads to notable gains in ACC, IC-LPIPS, and ARS, demonstrating its crucial role in improving mask-image alignment and synthesis quality.~\textbf{(3) Low-Quality Sample Filtering (LQSF):} The improved ACC with LQSF confirms that the proposed ARS metric effectively selects high-quality and well-aligned anomaly image–mask pairs.
\begin{table}[t]
    \renewcommand{\arraystretch}{0.9} 
    \setlength{\tabcolsep}{3pt} 
    \caption{Module ablation on MVTec-AD. “–” indicates stepwise removal of modules. In the “– RAMS” setting, ground-truth masks are randomly placed within the product's foreground region.}
    \begin{center}
        \resizebox{0.49\textwidth}{!}{
            \begin{tabular}{l|cccc}
                \toprule
                
                & IS & IC-LPIPS & ACC & ARS\\
                
                \midrule

                Ours(Full) & \textbf{1.89 }& \textbf{0.35} & \textbf{84.7} & \textbf{4.81}\\

                \midrule
                
                - AMFC & 1.82~\small{\color{red} ($\downarrow$ 0.07) } & 0.34~\small{\color{red} ($\downarrow$ 0.01) } & 81.3~\small{\color{red} ($\downarrow$ 3.4) } & 4.74~\small{\color{red} ($\downarrow$ 0.07)} \\

                - RAMS & 1.79~\small{\color{red} ($\downarrow$ 0.03) } & 0.31~\small{\color{red} ($\downarrow$ 0.03) } & 76.3~\small{\color{red} ($\downarrow$ 5.0) } & 4.58~\small{\color{red} ($\downarrow$ 0.16)}\\

                - LQSF & 1.78~\small{\color{red} ($\downarrow$ 0.01) } & 0.30~\small{\color{red} ($\downarrow$ 0.01) } & 73.6~\small{\color{red} ($\downarrow$ 2.7) } & 4.27~\small{\color{red} ($\downarrow$ 0.31)}\\
                \bottomrule
            \end{tabular}
        }
    \end{center}
     \vspace{-10pt}

    \label{tab:table_abalation}
\end{table}

\section{Conclusion and Limitation}
\label{sec:conclusion}
\noindent\textbf{Conclusion. }We propose GAA, a region-guided few-shot anomaly image–mask pair synthesis framework that generates realistic, diverse, and spatially aligned anomaly data using only a small number of anomaly samples. Built upon a pretrained Latent Diffusion Model, GAA introduces Localized Concept Decomposition and Adaptive Multi-Round Clustering strategies to jointly model the semantic features and spatial positions of anomalies. In addition, the proposed region-guided mask synthesis module enables precise control over anomaly regions, while the ARS-based sample filtering strategy further improves the quality and applicability of the synthesized dataset. Extensive experiments on the MVTec AD and MVTec LOCO datasets demonstrate that GAA significantly outperforms existing methods in both anomaly synthesis quality and downstream tasks such as localization and classification. This work provides a practical solution to the scarcity of anomaly data in industrial settings and lays a solid foundation for building controllable and scalable anomaly detection systems.

\noindent\textbf{Limitation.}~Although GAA has shown certain capabilities in generating logical anomalies, it still faces challenges in handling complex global layout anomalies. For example, relocating existing objects within an image or performing global transformations such as moving a transistor or flipping a metal nut remains difficult.

\bibliographystyle{ACM-Reference-Format}
\bibliography{main}







\clearpage
\appendix
\section{Overview}
In this supplementary material, we provide additional experimental details and result analyses.
\begin{itemize}
    \item \textcolor{red}{\S B.} More Implementation Details
    \item \textcolor{red}{\S C.} More Details of the Adaptive Anomaly Feature Cluster
    \item \textcolor{red}{\S D.} More Details of the Region-Guided Mask Generation
    \item \textcolor{red}{\S E.} Additional Experimental Results on MVTec AD
    \item \textcolor{red}{\S F.} Additional Experimental Results on MVTec LOCO
\end{itemize}

\section{More Implementation Details}
\subsection{Training of the Generation Model}
In our experiments, we assigned a feature-focused embedding $e_f$ to each new anomaly cluster to learn the feature representations of anomalies. Simultaneously, all anomaly clusters shared a common positional encoder $E$, which encodes the input anomaly mask $M_a$ into a position-focused embedding $e_p$ to capture spatial information. The positional encoder $E$ is composed of a ResNet-50 backbone, a Feature Pyramid Network (FPN), and fully connected layers. Specifically, ResNet-50 extracts features from the anomaly mask, the FPN fuses features from multiple layers, and the fully connected layers map the fused features into the embedding space.\\
The model was trained on an NVIDIA RTX 3090 24GB GPU with a batch size of 4 and an initial learning rate of $5 \times 10^{-3}$ for a total of 310,000 steps. During training, we utilized textual inversion by optimizing the pseudo-token ``\texttt{*}" to learn anomaly patterns. In our setup, each feature-focused embedding $e_f$ consists of 8 tokens, while each position-focused embedding $e_p$ consists of 4 tokens. On the MVTec AD dataset, the model was trained for over 310,000 steps, requiring approximately three days. On the MVTec LOCO dataset, the model was trained for 500,000 steps, taking approximately four days.
\subsection{Training of the Low-Quality Sample Filter}  
The automated data filter utilizes a fully connected architecture based on Simplenet with batch normalization for optimization. Gaussian noise $N(0, 0.015^2)$ is added to normal features to enhance data robustness. The Adam optimizer is employed, with learning rates of 0.0002 for the discriminator and 0.0001 for the feature adaptor, and a weight decay of 0.00001. The classification thresholds $th^+$ and $th^-$ are both set to 0.5. The model is trained on both the MVTec AD and MVTec LOCO datasets for 160 epochs on each dataset with a batch size of 4.
\subsection{Evaluation Metrics}
\begin{enumerate}
    \item \textbf{Inception Score (IS):} Evaluates the quality and diversity of generated images using KL divergence between their distributions and those predicted by the Inception model. Higher IS values indicate better quality and diversity.
    \item \textbf{IC-LPIPS Distance~(IC-L):} Measures intra-class consistency by assessing perceptual similarity between images within the same class. Higher IC-LPIPS scores suggest greater diversity.
    \item \textbf{Anomaly Region Score (ARS):} Evaluates the quality of generated anomaly-image pairs by calculating the pixel-level overlap between the anomaly score map and the anomaly mask. Higher ARS values indicate better alignment and quality.
    \item \textbf{AUROC:} Reflects classifier performance through the area under the Receiver Operating Characteristic (ROC) curve. Higher AUROC values denote better discrimination capability.
    \item \textbf{Average Precision (AP):} Measures classification performance via the area under the Precision-Recall curve. Higher AP values indicate better recognition, especially for minority classes.
    \item \textbf{$F_1$-max:} Represents the balance between precision and recall, defined as the maximum $F_1$ score across thresholds. Higher $F_1$-max scores indicate optimal performance.
    \item \textbf{Accuracy (ACC):} Represents the proportion of correctly predicted samples across the entire dataset. Higher ACC values denote better overall accuracy.
\end{enumerate}

\section{More Details of the Adaptive Anomaly Feature Cluster}
\label{sec:partC}
\begin{figure*}[h]
\centering
\includegraphics[width=0.95\textwidth]{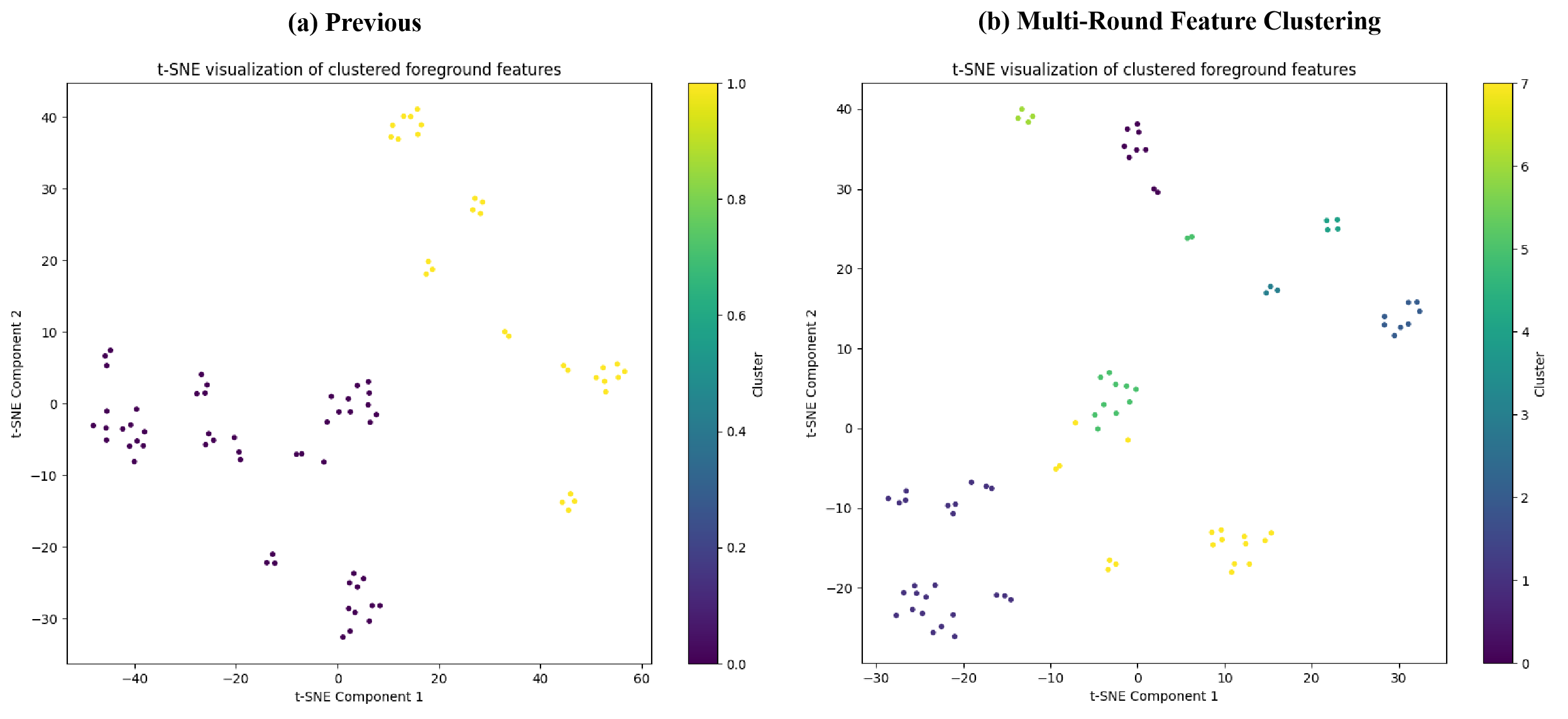}
\caption{Comparison of clustering results between the multi-round feature clustering method and the previous clustering method.}
\label{fig:cluster-cp}
\end{figure*}

\begin{figure*}[h]
\centering
\includegraphics[width=0.95\textwidth]{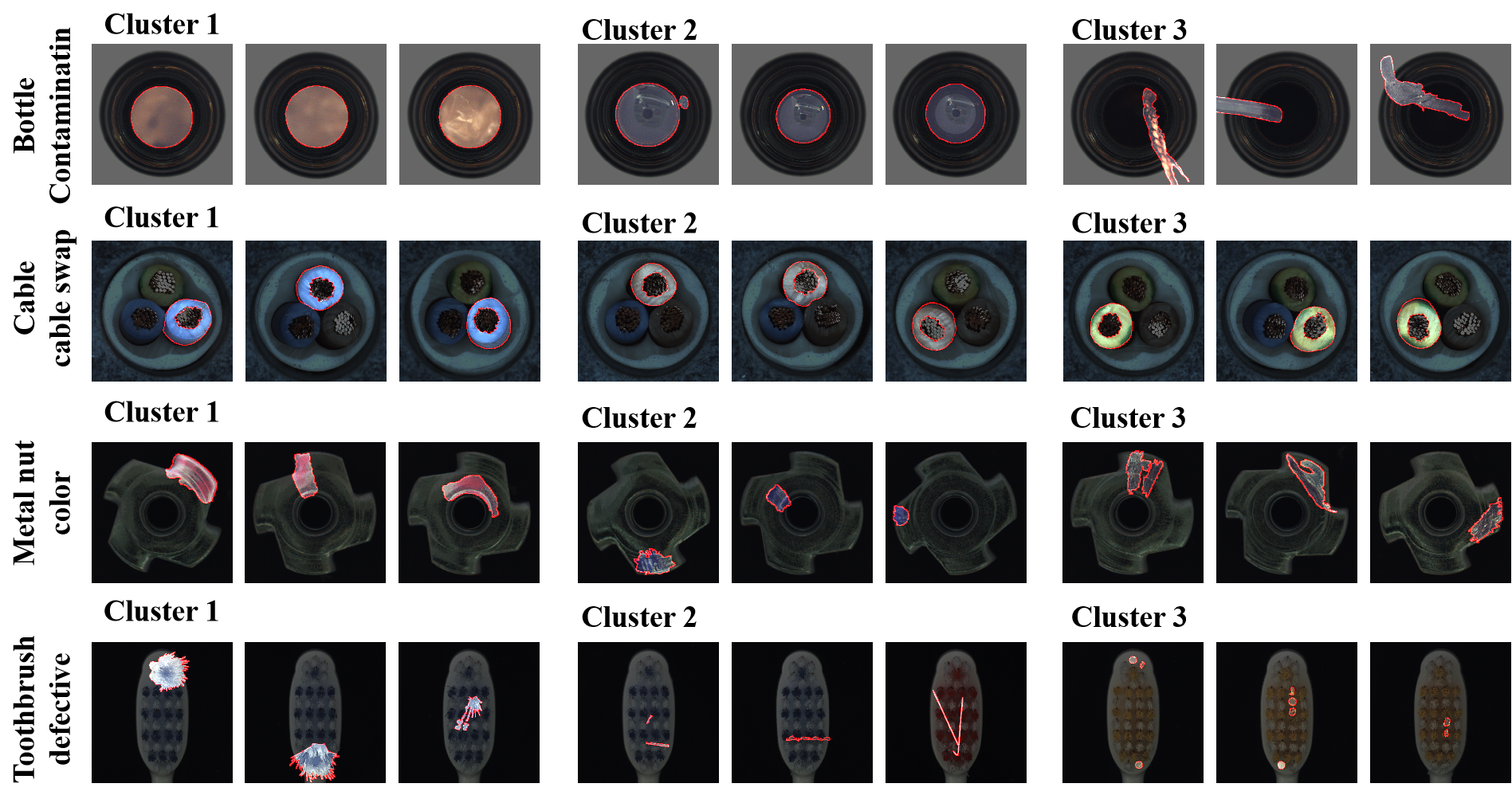}
\vspace{+5pt}
\caption{\textbf{Clustering results for specific anomaly categories in the MVTec AD dataset.} As shown in the figure, the multi-round clustering method effectively separates anomalies with different feature profiles. For example, in the ``cable" category, blue, green, and gray cables are grouped into distinct clusters. In the ``bottle" category, yellow contamination, transparent liquid-like contamination, and strip-shaped contamination are successfully separated into different clusters.}
\label{fig:AD-Cluster}
\vspace{+20pt}
\end{figure*}

\begin{figure*}[h]
\centering
\includegraphics[width=0.88\textwidth]{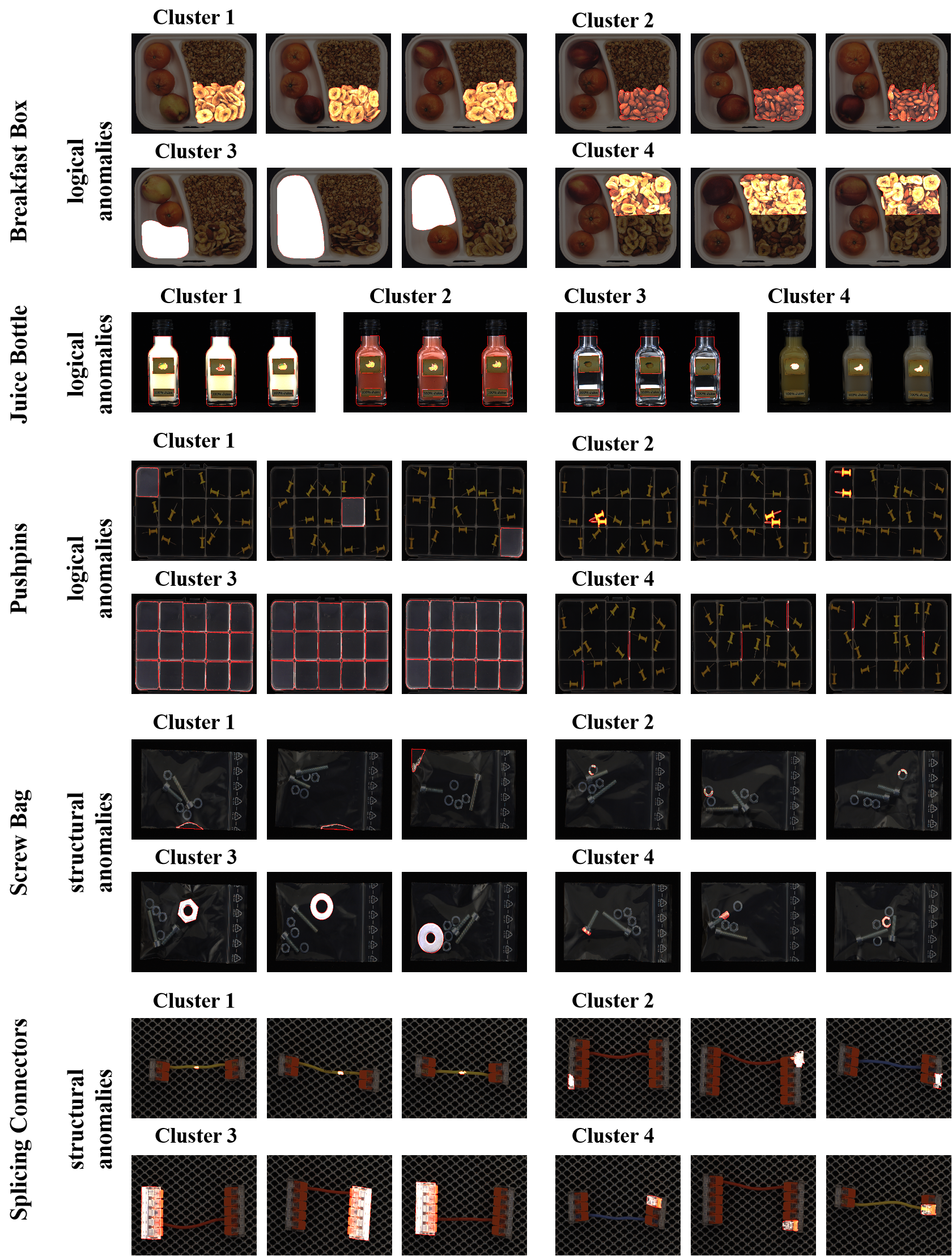}
\caption{\textbf{Clustering results for specific anomaly categories in the MVTec LOCO dataset.} As shown, the multi-round clustering method effectively separates anomalies with distinct feature profiles. For instance, in the ``breakfast box" category, anomalies such as only bananas, only nuts, and missing fruits are grouped into different clusters. In the ``juice bottle" category, red and yellow beverages, as well as empty bottles, are successfully separated into distinct clusters.}

\label{fig:LOCO-Cluster}
\end{figure*}

\textbf{Multi-round Feature Clustering:} Our approach utilizes a multi-round feature clustering method, which quantitatively evaluates both intra-cluster compactness and inter-cluster separability. This allows for the effective identification of anomalies with similar dense feature profiles, yet belonging to different categories, as illustrated in Figure~\ref{fig:cluster-cp}.\\
\noindent\textbf{Clustering Results:} The clustering results for specific anomaly categories in the MVTec AD and MVTec LOCO datasets are presented in Figure~\ref{fig:AD-Cluster} and Figure~\ref{fig:LOCO-Cluster}, respectively. For each category, we highlight 3 to 4 newly identified clusters.

%

\section{More Details about Region-Guided Aligned Mask Generation}
Our approach directs anomaly generation to specific target regions through mask localization, simulating real-world industrial scenarios by applying customized localization techniques for different anomaly types.

\begin{figure*}[!htbp]
\centering
\includegraphics[width=0.88\textwidth]{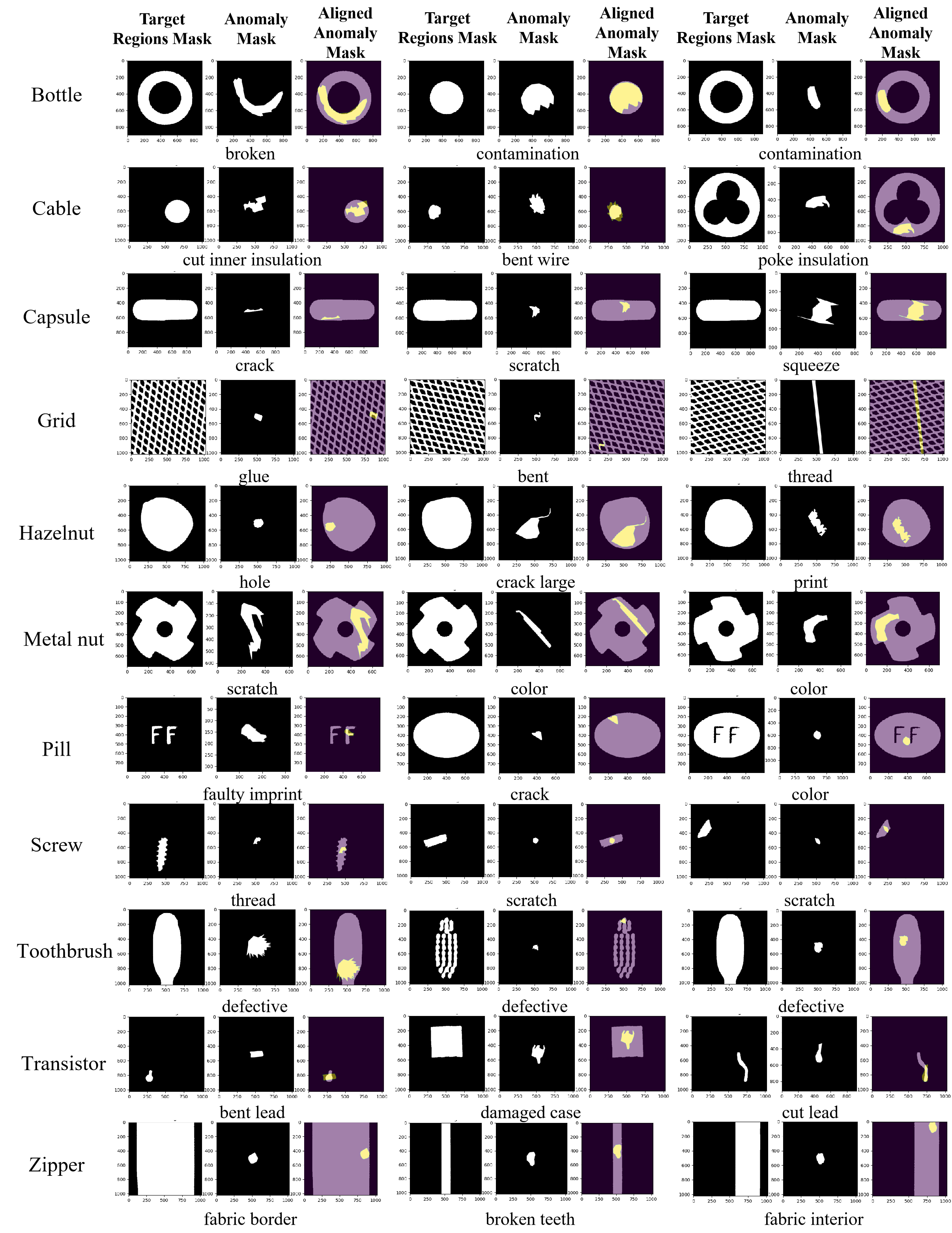}
\caption{\textbf{Region-guided aligned mask generation.} A large number of high-quality aligned anomaly masks for structural anomalies are synthesized using various localization methods, enabling diverse anomaly generation across different products.}
\label{fig:Region-Aligned}
\end{figure*}

\begin{itemize}
    \item \textbf{Structural Anomalies:} For structural anomalies such as broken, contamination, crack, and scratch, which can occur at various product locations, we randomize rotation and adjust mask size within target regions to diversify anomaly placements. For anomalies like bent wire in the wire region of a cable, we align the mask with the wire's center and apply a tolerable offset threshold to allow slight overflow, simulating realistic bending. Similar strategies are applied to anomalies like cut outer insulation in cables and fabric border anomalies in zippers by aligning the mask tangentially to the target region edges. These methods are demonstrated in Fig.~\ref{fig:Region-Aligned}, where region-guided masks reflect real-world anomaly distributions.

    \item \textbf{Logical Anomalies:} Logical anomalies, such as missing, misalignment, or mixing, generally involve changes in object presence or configuration. For these anomalies, the target region mask often serves directly as the anomaly mask. For example, in the breakfast box product, the nut region can simulate missing nuts or the absence of specific components like dried bananas. Similarly, in the cable product, the wire region can represent a missing wire anomaly or a cable swap anomaly, where insulation colors are interchanged.

    \item \textbf{Combined Anomalies:} For combined anomalies, we simulate the presence of multiple anomalies on a single good image. Masks for various anomalies, whether of the same or different types, are randomly selected while ensuring they do not overlap beyond a set threshold. This allows the creation of combined anomaly masks and the subsequent generation of diverse anomalies within the combined regions.
\end{itemize}

This customized augmentation strategy facilitates robust generalization across diverse anomaly types, addressing alignment challenges between anomalies and masks in industrial datasets. The results demonstrate high-quality anomaly generation and alignment, offering strong performance across numerous anomaly scenarios.

\section{More Generated Results on MvTec AD}
\begin{figure*}[!htbp]
\centering
\includegraphics[width=0.80\textwidth]{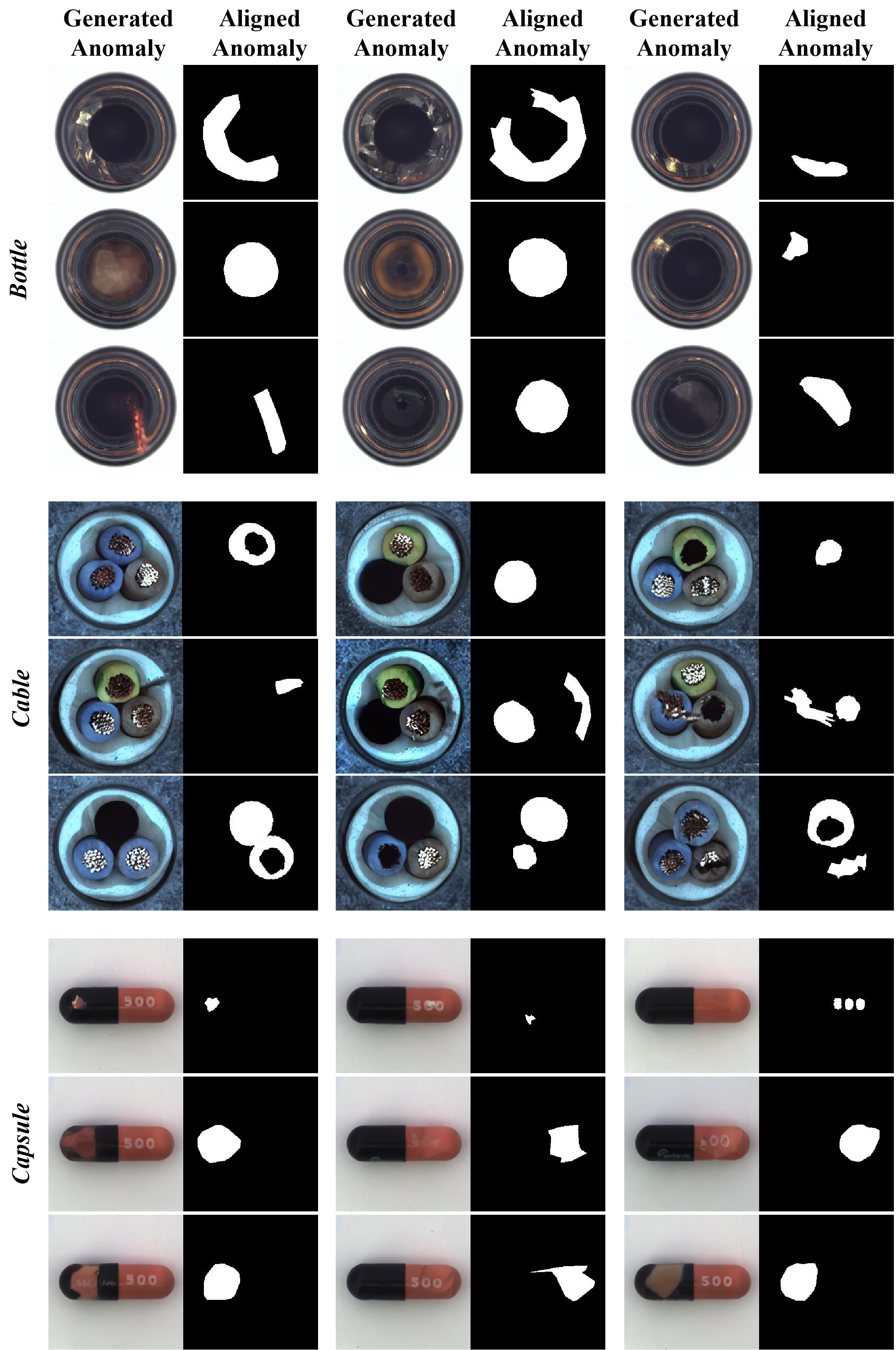}
\caption{Results generated on the MVTec AD dataset.}
\label{fig:Generated-GAA-AD-1}
\end{figure*}

\begin{figure*}[!htbp]
\centering
\includegraphics[width=0.80\textwidth]{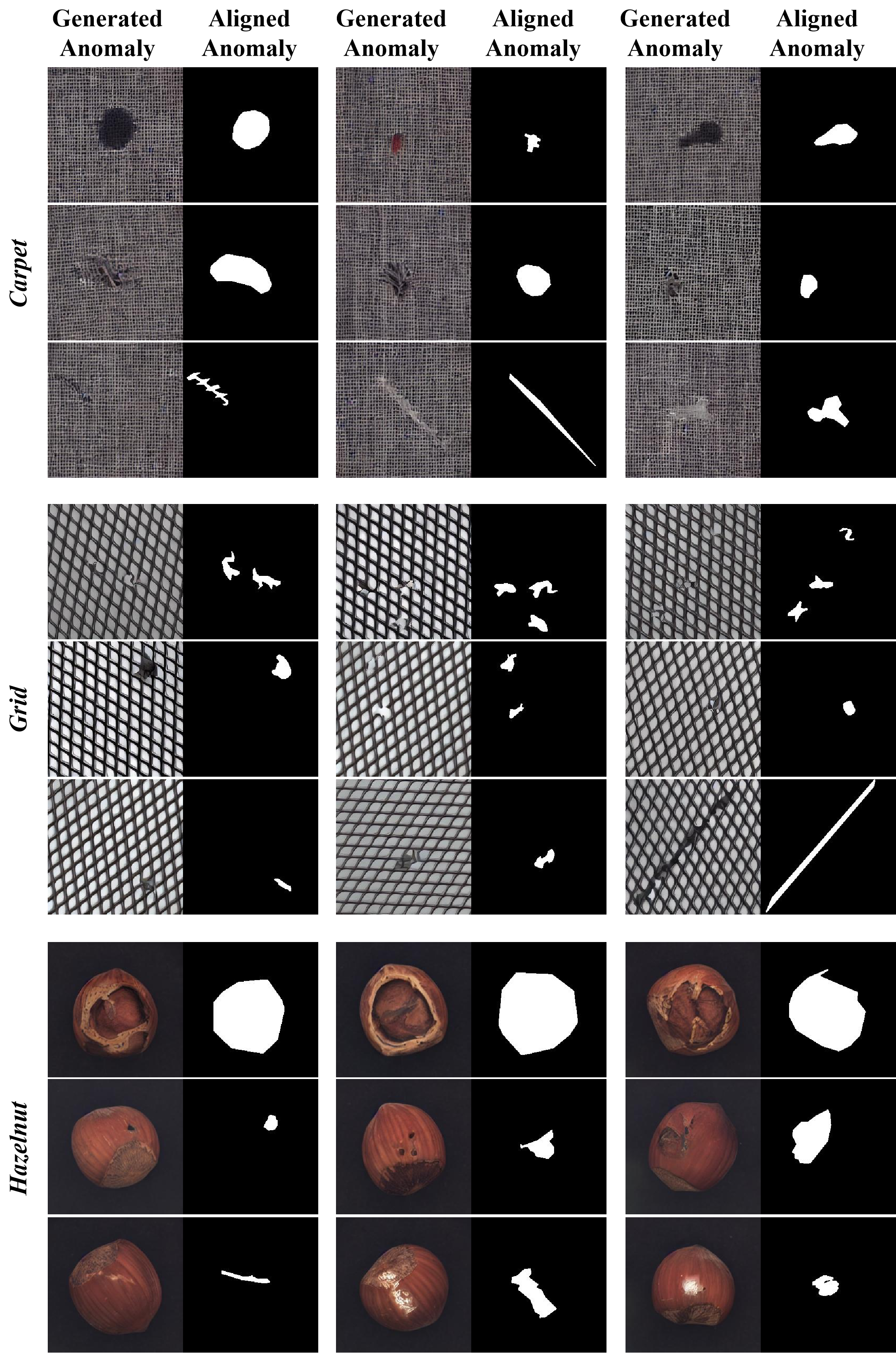}
\caption{Results generated on the MVTec AD dataset.}
\label{fig:Generated-GAA-AD-2}
\end{figure*}

\begin{figure*}[!htbp]
\centering
\includegraphics[width=0.80\textwidth]{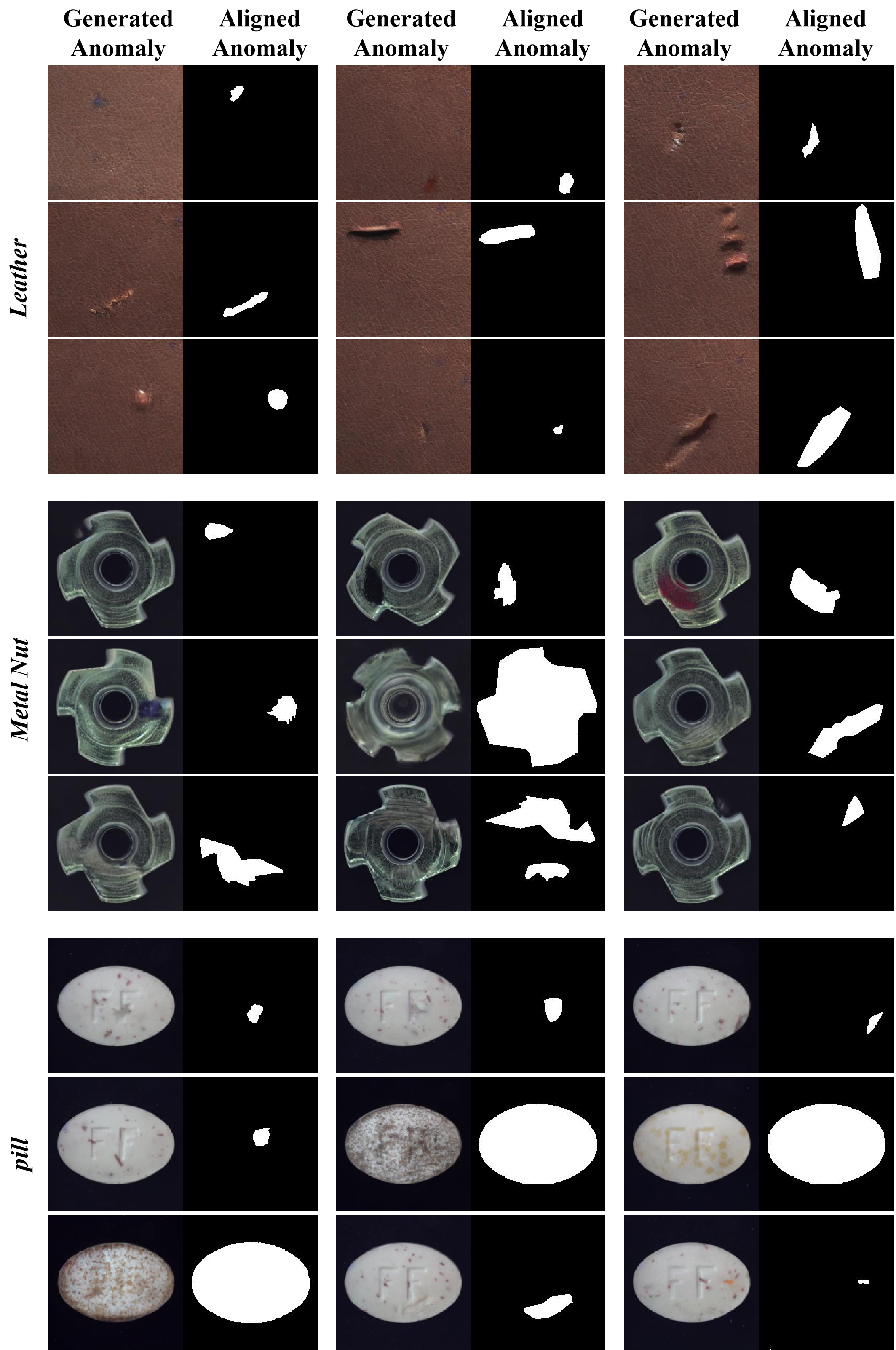}
\caption{Results generated on the MVTec AD dataset.}
\label{fig:Generated-GAA-AD-3}
\end{figure*}

\begin{figure*}[!htbp]
\centering
\includegraphics[width=0.80\textwidth]{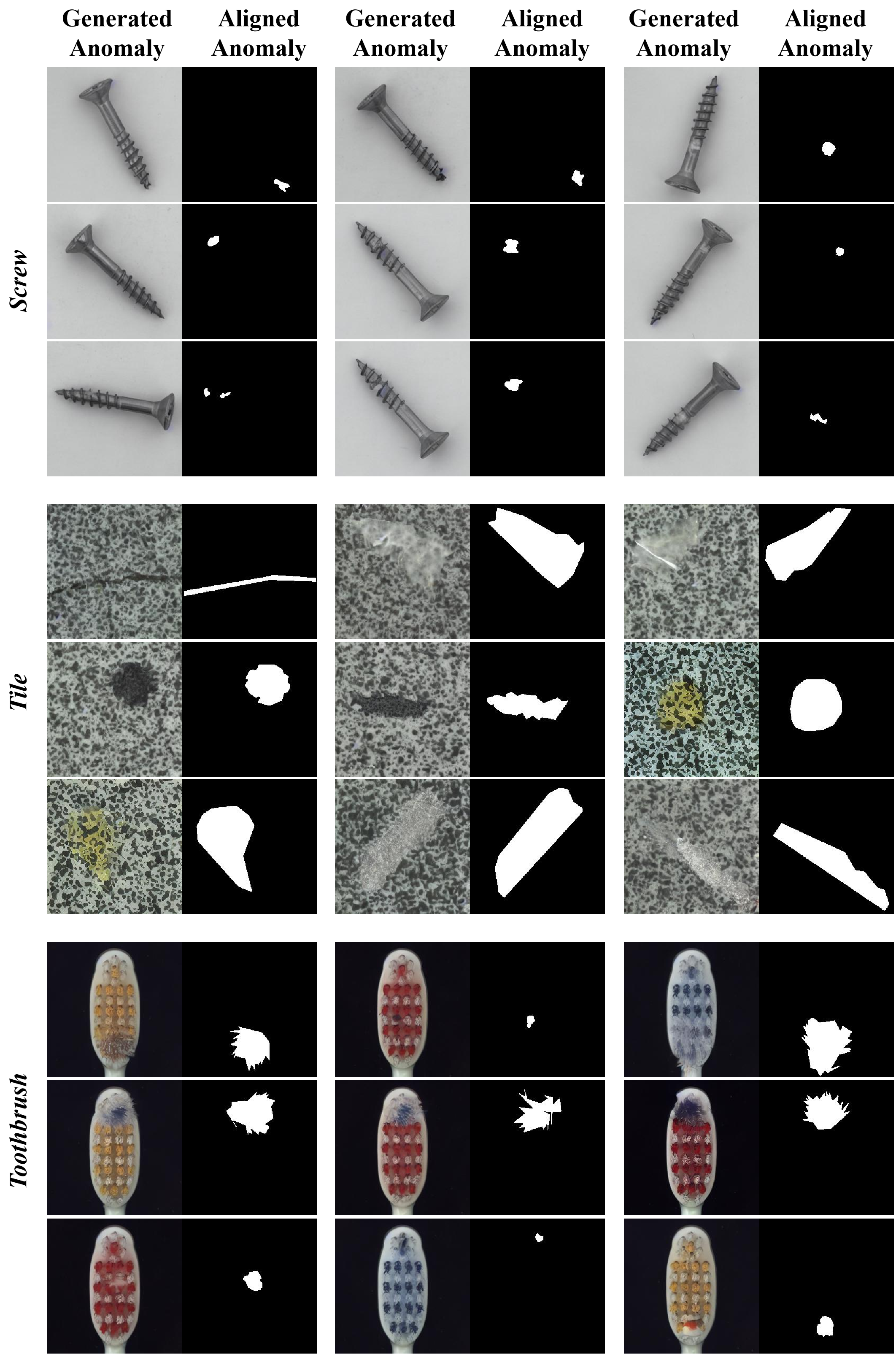}
\caption{Results generated on the MVTec AD dataset.}
\label{fig:Generated-GAA-AD-4}
\end{figure*}

\begin{figure*}[!htbp]
\centering
\includegraphics[width=0.80\textwidth]{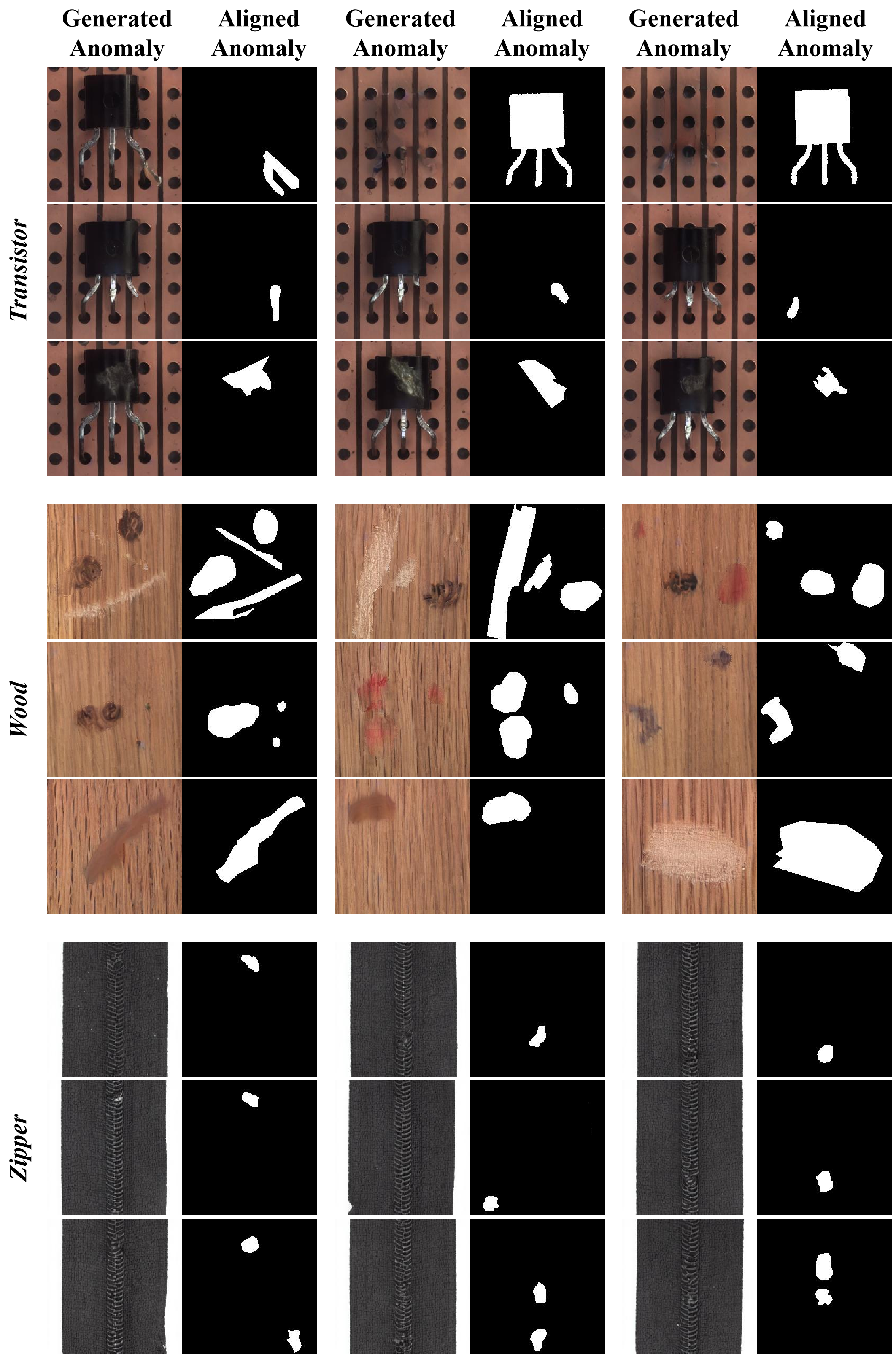}
\caption{Results generated on the MVTec AD dataset.}
\label{fig:Generated-GAA-AD-5}
\end{figure*}

In Figures~\ref{fig:Generated-GAA-AD-1},~\ref{fig:Generated-GAA-AD-2},~\ref{fig:Generated-GAA-AD-3},~\ref{fig:Generated-GAA-AD-4}, and~\ref{fig:Generated-GAA-AD-5} we sequentially provide a more comprehensive demonstration of the generated results. We generated anomalies with high fidelity, high alignment, high quality, and diversity across different types of anomalies for various products. Even when dealing with extremely challenging anomaly types, such as logical anomalies and composite anomalies, our method performs exceptionally well.
\section{More Generated Results on MvTec LOCO}
\begin{figure*}[!htbp]
\centering
\includegraphics[width=0.64\textwidth]{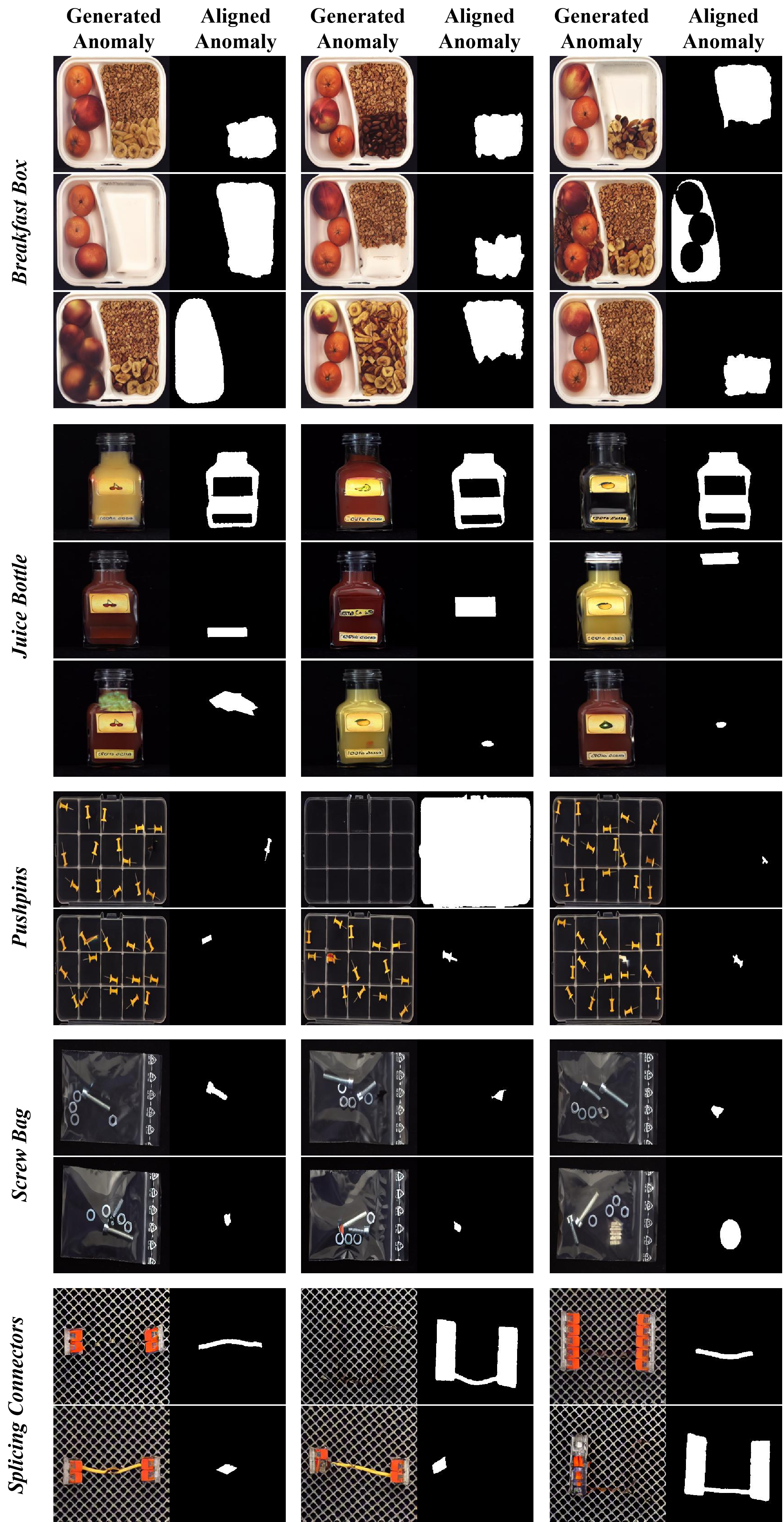}
\caption{Results generated on the MVTec LOCO dataset.}
\label{fig:Generated-GAA-LOCO}
\end{figure*}

In Figures~\ref{fig:Generated-GAA-LOCO}~,we sequentially provide a more comprehensive demonstration of the generated results. We generated anomalies with high fidelity, high alignment, high quality, and diversity across different types of anomalies for various products. Even when dealing with extremely challenging anomaly types, such as logical anomalies and multi-instance structural anomalies, our method performs exceptionally well.
\end{document}